\documentclass[letterpaper, 10 pt, conference]{ieeeconf}

\IEEEoverridecommandlockouts

\usepackage{pifont}

\usepackage{amsmath}
\usepackage{amssymb}
\usepackage{bm}
\usepackage{glossaries}
\usepackage{booktabs}
\usepackage{siunitx}
\usepackage{multirow}
\usepackage{siunitx}
\usepackage[plain]{fancyref}
\usepackage{cleveref}

\usepackage{balance}

\usepackage[labelformat=simple]{subcaption}
\DeclareCaptionLabelSeparator{periodspace}{.\quad}
\usepackage{hyperref}

\usepackage{tikz}
\usepackage{pgfplots}
\usetikzlibrary{patterns}
\usetikzlibrary{calc}
\usetikzlibrary{fit,positioning}

\usetikzlibrary{shapes,snakes}

\pgfplotsset{
    compat=newest,
    /pgfplots/legend image code/.code={%
        \draw[mark repeat=2,mark phase=2,#1] 
            plot coordinates {
                (0cm,0cm) 
                (0.5cm,0cm)
            };
    },
}

\definecolor{mycolor_light_green}{RGB}{155,255,155}
\definecolor{mycolor_light_orange}{RGB}{255,203,0}

\definecolor{mycolor1}{rgb}{1.00000,0.00000,0.00000}
\definecolor{mycolor2}{rgb}{0.00000,0.00000,1.00000}
\definecolor{mycolor3}{rgb}{0.00000,0.50000,0.00000}

\definecolor{mycolor4}{rgb}{1.00000,0.00000,0.00000}
\definecolor{mycolor6}{rgb}{0.00000,0.00000,1.00000}
\definecolor{mycolor5}{rgb}{0.00000,0.50000,0.00000}

\newcommand{\agent}{V}
\newcommand{\agentSet}{\mathcal{\agent}}
\newcommand{\agentMaxIndex}{K}

\newacronym{tum}{TUM}{Technische Universität München}
\newacronym{adas}{ADAS}{Advanced Driver Assistance System}
\newacronym{fsm}{FSM}{Finite State Machine}
\newacronym{hsm}{HSM}{Hierarchical State Machine}
\newacronym{mcdm}{MCDM}{Multiple Criteria Decision Making}
\newacronym{swrl}{SWRL}{Semantic Web Rule Language}
\newacronym{dl}{DL}{Description Logic}
\newacronym[longplural = {Terminological Boxes}, shortplural = {Tboxes}]{tbox}{TBox}{Terminological Box}
\newacronym[longplural = {Assertional Boxes}, shortplural = {Aboxes}]{abox}{ABox}{Assertional Box}
\newacronym[longplural = {Markov Decision Processes}]{mdp}{MDP}{Markov Decision Process}
\newacronym[longplural = {Partially Observable Markov Decision Processes}]{pomdp}{POMDP}{\textit{Partially Observable Markov Decision Process}}
\newacronym[longplural = {Mixed Observability Markov Decision Processes}]{momdp}{MOMDP}{Mixed Observability Markov Decision Process}
\newacronym[longplural = {Point-Based Markov Decision Processes}]{qmdp}{QMDP}{Point-Based Markov Decision Process}
\newacronym{dbn}{DBN}{Dynamic Bayesian Network}
\newacronym{hmm}{HMM}{Hidden Markov Model}
\newacronym{v2v}{V2V}{Vehicle-to-Vehicle}
\newacronym{v2i}{V2I}{Vehicle-to-Infrastructure}
\newacronym{abt}{ABT}{Adaptive Belief Tree}
\newacronym{tapir}{TAPIR}{\textit{Toolkit for approximating and Adapting POMDP solutions in Real time}}
\newacronym{ros}{ROS}{Robot Operating System}
\newacronym{oem}{OEM}{Original Equipment Manufacturer}
\newacronym{fdm}{FDM}{Foresighted Driver Model}
\newacronym{kit}{KIT}{Karlsruhe Institut für Technologie}
\newacronym{mpc}{MPC}{Model Predictive Control}
\newacronym{acc}{ACC}{Adaptive Cruise Control}
\newacronym{ctrv}{CTRV}{constant turn rate and velocity}
\newacronym{BN}{BN}{Bayesian network}
\newacronym{DBN}{DBN}{dynamic Bayesian network}
\newacronym{DBNs}{DBNs}{dynamic Bayesian networks}
\newacronym{IDM}{IDM}{Intelligent Driver Model}
\newacronym{HMM}{HMM}{hidden Markov model}
\newacronym{LSTM}{LSTM}{long short-term memory}

\makeglossaries

\newcommand\mytodo[1]{\textcolor{red}{@ToDo: #1}}
\renewcommand\mytodo[1]{}

\newcommand\jens[1]{\textcolor{orange}{Jens: #1}}

\title{\LARGE \bf 
Interaction-Aware Probabilistic Behavior Prediction in Urban Environments
}

\author{Jens Schulz$^{1}$, Constantin Hubmann$^{1}$, Julian L{\"o}chner$^{1}$, and Darius Burschka$^{2}$
\thanks{$^{1}$Jens Schulz, Constantin Hubmann, and Julian L{\"o}chner are with BMW Group, Munich, Germany
        {\tt\small \{jens.schulz $|$ constantin.hubmann $|$ julian.loechner\}@bmw.de}}%
\thanks{$^{2}$Darius Burschka is with the Department of Computer Science, Technical University of Munich, Germany
        {\tt\small burschka@tum.de}\quad\, \raisebox{0.2mm}{\textcopyright} 2018 IEEE}%
}

\begin{document}

\maketitle

\begin{abstract}

Planning for autonomous driving in complex, urban scenarios requires accurate prediction of the trajectories of surrounding traffic participants. Their future behavior depends on their route intentions, the road-geometry, traffic rules and mutual interaction, resulting in interdependencies between their trajectories.
We present a probabilistic prediction framework based on a dynamic Bayesian network, which represents the state of the complete scene including all agents and respects the aforementioned dependencies. We propose Markovian, context-dependent motion models to define the interaction-aware behavior of drivers.
At first, the state of the dynamic Bayesian network is estimated over time by tracking the single agents via sequential Monte Carlo inference. Secondly, we perform a probabilistic forward simulation of the network's estimated belief state to generate the different combinatorial scene developments. This provides the corresponding trajectories for the set of possible, future scenes.
Our framework can handle various road layouts and number of traffic participants. 
We evaluate the approach in online simulations and real-world scenarios. 
It is shown that our interaction-aware prediction outperforms interaction-unaware physics- and map-based approaches.

\mytodo{Advantage:
generalization to arbitrary number of possible routes and maneuvers by combination of Bayesian estimation and feature-based behavior models (can be learned from data).
Typically in literature: only learn discrete classifier for fix number of routes (e.g., right, left, straight, stop).}
\end{abstract}

\IEEEpeerreviewmaketitle

\section{Introduction}

\mytodo{refer more to classical robotics??}

While autonomous driving has already been pioneered in the 1980s by universities such as Carnegie Mellon and the Bundeswehr University Munich, it is still considered a challenge to integrate autonomous vehicles into real traffic.
A major difficulty is the interaction with human drivers.
Autonomous vehicles need to estimate the intentions and anticipate the future behavior of humans in order to plan collision-free trajectories and drive in a foresighted, efficient and cooperative manner. 
As intentions cannot be measured directly and humans exhibit individual and complex behavior, predictions will always be afflicted with uncertainty.

\begin{figure}
      \vspace{5px}
\centering
\includegraphics[scale=0.25, trim=0 180 0 177, clip=true]{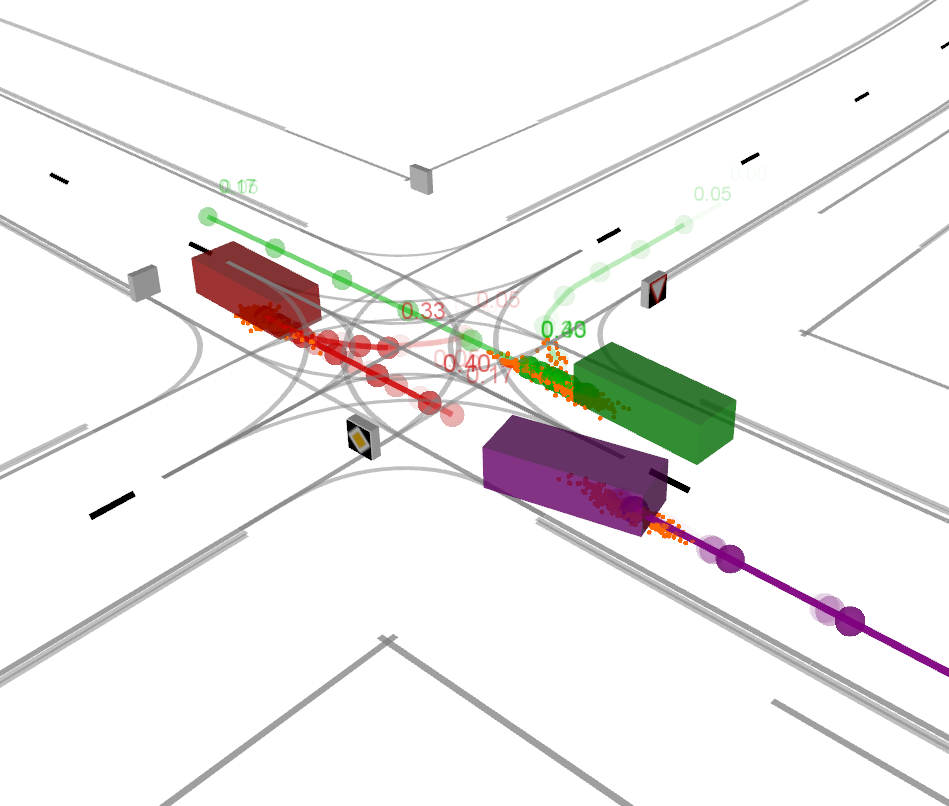}
\caption{Interaction-aware probabilistic trajectory prediction in an urban intersection scenario: the three vehicles have multiple possible routes, overlapping lanes and have to interact with each other.}
\label{fig:title}
\vspace{-9px}
\end{figure}

Simple prediction approaches such as constant turn rate and velocity may be sufficient for short term predictions and non-interactive situations. 
However, they quickly come to a limit in complex urban scenarios.
The mixture of crossing, merging and diverging lanes and corresponding traffic rules create a complex structure and a stronger need for interaction between traffic participants, as can be seen in \Fref{fig:title}:
the behavior of a driver depends on his intentions, the interactions with surrounding traffic and the static context, such as the road geometry.
Furthermore, the future trajectory of a vehicle does also depend on how the complete situation evolves over time and, therefore, on how other agents are going to act.
This introduces the need for combinatorial and interaction-aware motion prediction, which still represents a great challenge today \cite{lefevre_survey_2014}.

\mytodo{
As shown with the frozen robot problem \cite{trautman_unfreezing_2010}, to solve such complex environments the interaction have to be considered in planning and prediction
}

\mytodo{
Generalizable to arbitrary number of routes (not only classification of left right straight).
Can be used as a simulator for verification
}

In this paper, a behavior prediction framework is presented, which explicitly considers the intentions of drivers and the interdependencies between their future trajectories.
We model the development of a traffic situation as a stochastic process consisting of multiple interacting agents.
The decision making process of an agent is divided into three hierarchical layers: which route it is going to follow (route intention),
whether it is going to pass a conflict area at an intersection before or after another agent
(maneuver intention), and what continuous action it is going to execute.
First, the set of possible routes and maneuvers is queried online given a digital map and the agents' poses. 
Each agent then acts according to context-dependent behavior models given their route and maneuver intentions and the current environment.
Describing this process as a \gls{DBN} allows to specify causal as well as temporal dependencies and consider uncertainty in measurements and human behavior. Sequential Monte Carlo inference, also known as particle filtering, enables the use of hybrid, non-linear system models and the representation of arbitrary probability distributions.
Using observations of the agents' poses and velocities, Bayesian statistics allow for an estimation of the intentions and, therefore, for a more accurate probabilistic trajectory prediction by forward simulation of the \gls{DBN}.

In this work, we focus on unsignalized intersections due to the prevalence of interdependencies between vehicles.

\mytodo{
Why so complicated / interaction awareness?
\begin{itemize}
\item simple approaches exist (e.g. CTRV, neglecting interactions etc.), highway only, no interaction at intersections
\item cite here already?
\item complex scenarios, interactions
\item multi-agent environments
\item high variance in human actions, therefore probabilistic models
\item different road layouts (merge, intersect, etc.), T-junctions, arbitrary number of options
\item generalization
\item combine planning and machine learning
\item maintain interpretability
\end{itemize}
}

\mytodo{from Consti:
Why do we do this research:
\begin{itemize}
\item extreme development race in the area of autonomous driving
\item acceptability of such systems also based on comfort (motion sickness problems), comprehensible behavior and a safe feeling which is different to absolute safety which is possible without prediction (e.g. standstill) but delivers very suboptimal behavior
\item \jens{@Constantin: I wouldn't say prediction is completely unnecessary for safety per se, I would say a FANCY prediction (considering interactions, intentions and things like these) is not necessary, but any motion planning algorithm needs an estimate (no matter how bad it is) of the motion of others. Constant velocity is also prediction!}
\item therefore prediction of other traffic participants absolutely necessary to fulfill the requirements in the item above
\item prediction uncertainty arises from:
\begin{itemize}
\item measurement noise
\item unknown intention (e.g. goal)
\item unknown driving style (e.g. longitudinal behavior)
\end{itemize}
\item Highway:
\begin{itemize}
\item prediction of lane changes sufficient (only three classes)
\item longitudinal behavior and safety can absolutely done by control algorithm
\item considering the vehicle's state alone is enough
\end{itemize}
\item City: 
\begin{itemize}
\item mixture of crossing lanes, split roads and merging lanes
\item considering the state of the single, predicted vehicle not enough
\item tracking over time promises better result
\end{itemize}
\end{itemize}
Interaction between dynamic obstacles must be considered as shown by \cite{trautman_unfreezing_2010} to provide realistic behavior or prediction.
}

\mytodo{from Consti:
While autonomous driving was pioneered by military research competitions like the DARPA challenge, an extreme race arose over the last years between various companies to bring the first autonomous cars on the road. Considered as a game changer for the mobility industry, car companies suddenly have to compete with start-ups as well as software companies.

The proven safety of such robotic systems is named as the crucial factor to bring autonomous cars to the market. While this is surely the case, the perceived comfort and comprehensibility of the driving style may be evenly important to increase the acceptance of the customers. To allow motion planning algorithms like \gls{mpc} to plan trajectories, a prediction of the future movement of the surrounding traffic is absolutely needed. While simple prediction approaches as constant velocity may even lead to safe systems, the behavior can be very suboptimal leading to motions with high jerk when the prediction changes. 

As the development of autonomous driving started within the structured case of highways, classification of simple lane change maneuvers via supervised-learning is often used for such cases. As the longitudinal behavior may be realized without complex prediction algorithms, this type of prediction is sufficient for must \gls{acc} cases.
While it may be sufficient, to even only choose the leading vehicle as target object for driving on highways, urban environments require situation interpretation and prediction of the whole scene. This is the case as the inner-cities' mixture of crossing, split and merging lanes creates a more complex structure. As shown with the frozen robot problem \cite{trautman_unfreezing_2010}, to solve such complex environments the interaction have to be considered in planning and prediction (@JENS: Noch nicht wirklich gut,ich weiss. Ich will darauf hinaus dass man die ganzen inferierenden Einfluesse modellieren muss um plannen und/oder praedizieren zu koennen).
}

\mytodo{
Main contribution: particle-based state estimation and prediction framework
\begin{itemize}
\item automated intention definition by domain knowledge based on physical constraints and traffic rules
\item feature generation by domain knowledge
\item particle-based probabilistic calls to deterministic decision-making library
\item dummy library based on simple rules (IDM, etc.)
\end{itemize}
}

\mytodo{
Properties:
\begin{itemize}
\item arbitrary probabilistic forward simulation
\item model of dependencies based on state only, therefore no need for global solution or iterative solution
\item neglecting interdependencies ("if he does A, then I do B")
\item mild assumption? with machine learning this could be learned implicitly
\item basically just feature generation / transformation / processing
\end{itemize}
}

\mytodo{
To realize this, we present an 
\begin{itemize}
\item tracking based algorithm, which respects measurements of different time steps in the state-space estimation
\item Model scene development as stochastic process
\item consisting of decision making processes of all traffic participants
\item arbitrary number of traffic participants
\item main contribution: extension of IDM to be probabilistic (sampling within range and based on utility), extension to curvature, traffic lights (other velocity implications), crossings
\item ...
\item present probabilistic goal/route mapping and longitudinal behavior
\end{itemize}
One of our keyaspects is that we calculate the most probable motion homotopies for the other vehicles analytically. This simplification allows fast calculation while presenting respecting the different motion hypotheses.
}

\mytodo{
For an autonomous vehicle within an urban environment, it is important to gain a scene understanding and predict how a traffic situation as a whole is going to develop over time. 
The motion of multiple traffic participants is often highly interdependent and influences by a large set of features. Therefore ...

The decision making process of an agent is divided into different hierarchical parts, which are described in detail in the following subsections. 
}
\mytodo{shortly explain the purpose of the single nodes and then reference the figure}
\mytodo{
WHAT WE DO
\begin{itemize}
\item consider each traffic participant to be an agent, that can choose an action to influence its motion
\item predict whole scene development
\item consisting of multiple agents (vehicles, etc.)
\item each agent can conduct actions
\item actions depend on environment and intentions
\item goal driven, intentions
\item context dependent
\item assumption: future trajectories of multiple agents are not directly interdependent, but only depend on current multi-agent state (might implicitly be interdependent).
\item assumption: action only depends on past and current time (not implemented yet, but agents might also try to estimate intentions of others)
\item model these decision making processes as stochastic process within a DBN
\item infinite forward simulation possible, basically a traffic simulator with multiple hypotheses of intentions
\item FIGURE OF NETWORK (explain one time slice and forward dependency between time steps)
\end{itemize}
}

\section{Related Work}

In the area of autonomous vehicles, intention estimation and motion prediction of traffic participants has been widely studied.
Although these problems are highly coupled, in the existing literature, they are often tackled separately.

\subsection{Intention Estimation}
Popular methods for estimating route and maneuver intentions are discriminative classifiers (e.g., support vector machines (SVMs) \cite{aoude_behavior_2011}, random forests (RFs) \cite{barbier_classification_2017}, artificial neural networks (ANNs) \cite{phillips_generalizable_2017}) and probabilistic graphical models (e.g., hidden Markov models (HMMs) \cite{streubel_prediction_2014}, Bayesian networks (BNs) \cite{liebner_driver_2012}).
For this purpose, the set of possible intentions is typically predefined offline and the models are learned for these fixed number of classes.
For highways, this set usually consists of \emph{lane change left, lane change right,} and \emph{keep lane} (e.g., \cite{kumar_learning-based_2013, bahram_combined_2016}).
For intersections, the desired route is mostly represented by the turning directions \emph{left, right,} and \emph{straight} (e.g., \cite{streubel_prediction_2014, liebner_driver_2012}).
Besides the intention of a lane change or the desired route, more detailed intentions can be distinguished. A longitudinal classification whether to yield or stop before an intersection has already been investigated (e.g., \cite{aoude_behavior_2011, klingelschmitt_combining_2014, barbier_classification_2017}).
In \cite{lefevre_risk_2012}, the set of possible intentions is generate online: the possible route alternatives and the corresponding yield positions are determined online using a map and the intentions are estimated.

Interactions between traffic participants are often not considered (e.g., \cite{kumar_learning-based_2013, streubel_prediction_2014, aoude_behavior_2011, barbier_classification_2017}).
When the motion of multiple vehicles is interdependent, however, this may result in inaccurate predictions, especially for longer prediction horizons (e.g., if a vehicle approaching an intersection has to decelerate because of a slow vehicle in front, without considering interactions, it might be misleadingly inferred that it intends to turn).
Investigating the so-called \emph{freezing robot problem}, \cite{trautman_unfreezing_2010} has shown that agents typically engage in \emph{joint collision avoidance} and cooperatively make room to create feasible trajectories.
Therefore, possible future interactions between agents should be taken into account.

Others works on intention estimation have already explicitly modeled interdependencies between vehicles:
In \cite{liebner_driver_2012} and \cite{klingelschmitt_combining_2014}, the dependency on the preceding vehicle is considered in order to improve the estimation at intersections.
In \cite{kuhnt_understanding_2016}, interdependencies between multiple vehicles are modeled using object oriented probabilistic relational models with learned probability tables. They automatically extract the possible routes from the map and distinguish different interaction types depending on the route relations (\emph{merge, cross, diverge, follow}) of vehicles.
Promising results are shown by \cite{phillips_generalizable_2017} with a \gls{LSTM} based route classification for intersections, considering the states of up to seven surrounding vehicles, therefore, respecting possible interactions implicitly.

All of these works focus on intention estimation with discrete classes, but do not predict continuous trajectories needed for many motion planning algorithms.

\subsection{Trajectory Prediction}

The most simple trajectory prediction methods are physics-based and assume models like constant velocity, not considering the situational context \cite{lefevre_survey_2014}. Especially at intersections and for long prediction horizons, these models tend to have low accuracy due to the high dependency of the drivers' actions on the road geometry, traffic rules and interactions to surrounding vehicles.
Trajectory prediction that incorporates contextual information is often based on regression methods (e.g., Gaussian processes (GPs) \cite{tran_online_2014, armand_modelling_2013}, RFs \cite{gindele_learning_2013}, ANNs \cite{lenz_deep_2017}) or planning-based methods (e.g., \cite{schulz_estimation_2017}).
In \cite{armand_modelling_2013}, velocity profiles with heteroscedastic variance for stopping at an intersection are learned using GPs. They include knowledge about the upcoming intersection, but do not consider other vehicles.
In \cite{wheeler_analysis_2016}, seven different regression methods for interaction-aware microscopic driver behavior are learned and compared to each other in highway scenarios.
An ANN based mapping from Markovian scene state to a continuous action distribution of an agent is learned for highway scenarios by \cite{lenz_deep_2017}.
These models allow an interaction-aware forward simulation, but do not explicitly infer route or maneuver intentions.

\subsection{Estimation and Prediction}
Besides the work that is either concerned about intention estimation or trajectory prediction, there has been effort to solve these problems together:
A two-staged approach is employed by \cite{platho_predicting_2013}, in which they first classify a traffic situation into one of multiple predefined driving situations and then predict the velocity profile using situation-specific models. As these profiles only depend on features of the current situation (e.g., states of preceding vehicles), but do not incorporate the prediction of the surrounding vehicles, future interdependencies are ignored.
Another combined approach can be found in \cite{bahram_combined_2016}, where highway maneuvers are first estimated based on multi-agent simulations and then used as input for a continuous trajectory prediction.
Thus, both works solve the two problems separately, but improve their trajectory prediction by their maneuver and route estimates.

In \cite{gindele_learning_2013}, learned context-dependent action models of traffic participants are embedded into a DBN in order to estimate the state of the current situation applying sequential Monte Carlo (SMC) inference and predict the future motion of drivers. They outperform a Bayesian filter with constant velocity and heading assumption in simulations in terms of position accuracy. Although the different route options are modeled within the DBN, driver intentions are not explicitly inferred and evaluated.
GP regression is utilized by \cite{tran_online_2014} to estimate the predefined route intention also using SMC.
In our previous work \cite{schulz_estimation_2017}, we address the interrelated problems of behavior generation of the ego vehicle and behavior prediction of the surrounding vehicles in a combined fashion. Multi-agent maneuvers based on the concept of homotopy and corresponding trajectories are planned and used for intention estimation and ego vehicle control.

In contrast to the work presented in this section, we aim to propose a model for combined intention estimation and state prediction that can handle
\begin{itemize}
\item automatic generation of route hypotheses and maneuver hypotheses given the map and agent poses
\item a varying number of traffic participants and various intersection layouts
\item uncertainty in both measurements and human behavior
\item combinatoric interaction between traffic participants.
\end{itemize}

\mytodo{
\cite{petrich_assessing_2014}
\cite{petrich_map-based_2013}
\cite{tran_probabilistic_2012}
\cite{tran_online_2014}

ICRA und IROS paper!! RSS paper!
}

\section{Problem Statement}\label{sec:problem_statement}

A traffic scene $S$ consists of a set of agents ${\agentSet=\{\agent^0,\cdots\!,\agent^\agentMaxIndex\}}$, with $\agentMaxIndex \in \mathbb{N}_0$, in a static environment ($\mathrm{map}$) with discrete time, continuous state, and continuous action space. 
The map consists of a road network with topological, geometric and infrastructure (yield lines, traffic signs, etc.) information as well as the prevailing traffic rules.
At time step $t$, the set of agents $\agentSet$ is represented by their kinematic states ${X_t=[\bm{x}^0_t, \cdots\!, \bm{x}^\agentMaxIndex_t]^\top}$, route intentions ${R_t=[r^0_t,\cdots\!,r^\agentMaxIndex_t]^\top}$, and maneuver intentions ${M_t=[m^0_t,\cdots\!,m^\agentMaxIndex_t]^\top}$.
The kinematic state ${\bm{x}^i_t = [x^i_t,y^i_t,\theta^i_t,v^i_t]^\top}$ of agent $\agent^i$ consists of the Cartesian position, heading, and absolute velocity. 
Its length and width are considered to be given deterministically by the most recent measurement and, for the sake of brevity, are not included within $\bm{x}^i$.
The route intention $r^i_t$ defines a path through the road network the agent desires to follow, the maneuver intention $m^i_t$ the desired order relative to other agents in cases of intersecting or merging routes (see Sec. \ref{sec:sub_route_intention} and \ref{sec:sub_maneuver_intention} for detailed definitions). Other types of maneuvers such as lane changes or overtaking are not considered within this work.
At each time step, each agent executes an action $\bm{a}^i_t$ that depends on its intentions, the map and the kinematic states of all agents, transforming the current kinematic state $\bm{x}^i_t$ to a new state $\bm{x}^i_{t+1}$. The actions of all agents are denoted as $A=[\bm{a}^0_t,\cdots,\bm{a}^K_t]^\top$.
The complete dynamic part of a scene is thus described by ${S_t=[X_t,R_t,M_t,A_t]^\top}$.
At each time step, a noisy measurement ${Z_t=[\bm{z}^0_t,\cdots\!,\bm{z}^K_t]^\top}$ with ${\bm{z}^i_t = [z_{x,t}^i,z_{y,t}^i,z_{\theta,t}^i,z_{v,t}^i]^\top}$ is observed according to the distribution ${P(Z_t|X_t)}$, that contains information about the kinematic states of all agents.

The objective of this work is twofold: one part is to estimate the route intentions $R$ and maneuver intentions $M$ of all agents at the current time.
The other part is to predict the future kinematic states ${X}$ up to a temporal horizon $T$.

\mytodo{
given input: 
-measurements
-map
ground truth:
-future route R
-future vehicle states
desired output: 
-intention distribution: max  P(R)
-vehicle state prediction: min eukledean distance / maximum likelihood of ground truth given predicted distribution 
}

\section{Approach}\label{sec:approach}
\newlength{\timeslicedist}

\setlength{\timeslicedist}{1.3cm}
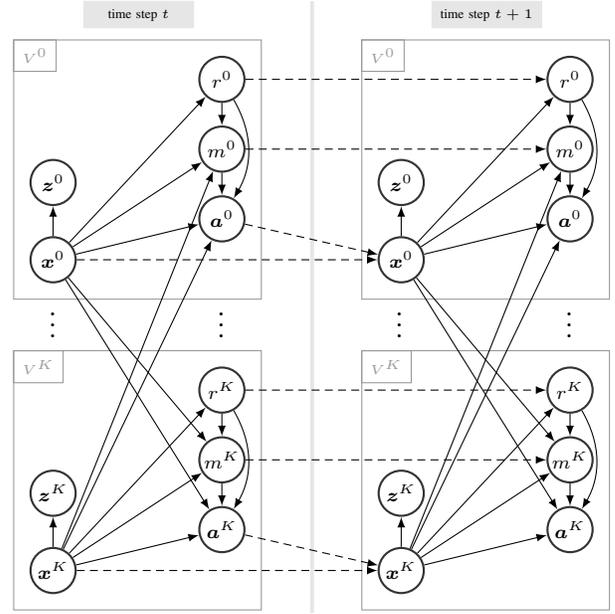
\begin{figure}
      \vspace{3px}
\centering
\scalebox{1}{
\begin{tikzpicture}
\tikzstyle{node_round}=[circle, minimum size =6mm, thick, draw =black!80, node distance = 5mm]
\tikzstyle{node_square}=[rectangle, minimum size = 6mm, thick, draw =black!80, node distance = 5mm]
\tikzstyle{connect}=[-latex]
\tikzstyle{box}=[rectangle, draw=black!100]

\draw[black!10, line width=0.5mm] (1.2cm,1.04) --(1.2cm,-7.1);

  \node[draw, black!40] at (-2.486cm,0.316) {\tiny \centering $\agent^0$};
  \node[draw, black!40] at (2.14cm,0.316) {\tiny \centering $\agent^0$};
  \node[draw, black!40] at (-2.44cm,-3.824) {\tiny \centering $\agent^\agentMaxIndex$};
  \node[draw, black!40] at (2.188cm,-3.824) {\tiny \centering $\agent^\agentMaxIndex$};

  \node[node_round,fill = black!0] (routeA1) [label=center:{\scriptsize $r^0$}] {};
  \node[node_round,fill = black!0] (maneuverA1) [below=0.3cm of routeA1,label=center:{\scriptsize $m^0$}] {};
  \node[node_round,fill = black!0] (actionA1) [below=0.3cm of maneuverA1,label=center:{\scriptsize $\bm{a}^0$}] {};
  \node[node_round, fill = black!0] (stateA1) [below left=0.1cm and 1.8cm of actionA1,label=center:{\scriptsize $\bm{x}^0$}] {};
  \node[node_round, fill = black!0] (measurementA1) [above=0.4cm of stateA1,label=center:{\scriptsize $\bm{z}^0$}] {};
  \node[rectangle, inner sep=2.1mm,draw=black!40,fill=black!30, fill opacity=0.0, fit= (routeA1) (maneuverA1) (actionA1) (stateA1)] {};

  \coordinate (stateA1_a) at ($(stateA1)+(0.3,0)$);
  \coordinate (routeA1_b) at ($(routeA1)-(0.3,0)$);
  \coordinate (routeA1_a) at ($(routeA1)+(0.3,0)$);
  \coordinate (maneuverA1_b) at ($(maneuverA1)-(0.3,0)$);
  \coordinate (maneuverA1_a) at ($(maneuverA1)+(0.3,0)$);
  \coordinate (actionA1_b) at ($(actionA1)-(0.3,0)$);
  \coordinate (actionA1_a) at ($(actionA1)+(0.3,0)$);

  \path (routeA1) edge [connect] (maneuverA1);
  \path (maneuverA1) edge [connect] (actionA1); 
  \path (routeA1) edge [bend left, connect] (actionA1);
  \path (stateA1) edge [connect] (routeA1);
  \path (stateA1) edge [connect] (maneuverA1);
  \path (stateA1) edge [connect] (actionA1);
  \path (stateA1) edge [connect] (measurementA1);

  \node[node_round,fill = black!0] (routeA2) [below=1.65cm of actionA1, label=center:{\scriptsize $r^K$}] {};
  \node[node_round,fill = black!0] (maneuverA2) [below=0.3cm of routeA2,label=center:{\scriptsize $m^K$}] {};
  \node[node_round,fill = black!0] (actionA2) [below=0.3cm of maneuverA2,label=center:{\scriptsize $\bm{a}^K$}] {};
  \node[node_round,fill = black!0] (stateA2) [below left=0.1cm and 1.8cm of actionA2,label=center:{\scriptsize $\bm{x}^K$}] {};
  \node[node_round, fill = black!0] (measurementA2) [above=0.4cm of stateA2,label=center:{\scriptsize $\bm{z}^K$}] {};
  \node[rectangle, inner sep=2.1mm,draw=black!40,fill=black!30, fill opacity=0.0, fit= (routeA2) (maneuverA2) (actionA2) (stateA2)] {};

  \coordinate (stateA2_a) at ($(stateA2)+(0.3,0)$);
  \coordinate (stateA2_b) at ($(stateA2)-(0.3,0)$);
  \coordinate (routeA2_b) at ($(routeA2)-(0.3,0)$);
  \coordinate (routeA2_a) at ($(routeA2)+(0.3,0)$);
  \coordinate (maneuverA2_b) at ($(maneuverA2)-(0.3,0)$);
  \coordinate (maneuverA2_a) at ($(maneuverA2)+(0.3,0)$);
  \coordinate (actionA2_b) at ($(actionA2)-(0.3,0)$);
  \coordinate (actionA2_a) at ($(actionA2)+(0.3,0)$);

  \path (routeA2) edge [connect] (maneuverA2);
  \path (maneuverA2) edge [connect] (actionA2);
  \path (routeA2) edge [bend left, connect] (actionA2);
  \path (stateA2) edge [connect] (routeA2);
  \path (stateA2) edge [connect] (maneuverA2);
  \path (stateA2) edge [connect] (actionA2);
  \path (stateA2) edge [connect] (measurementA2);

  \path (stateA1) edge [connect] (maneuverA2);
  \path (stateA1) edge [connect] (actionA2);
  
  \path (stateA2) edge [connect] (maneuverA1);
  \path (stateA2) edge [connect] (actionA1);

    \node[node_round,fill = black!0] (route2A1) [right=4cm of routeA1, label=center:{\scriptsize $r^0$}] {};
  \node[node_round,fill = black!0] (maneuver2A1) [below=0.3cm of route2A1,label=center:{\scriptsize $m^0$}] {};
  \node[node_round,fill = black!0] (action2A1) [below=0.3cm of maneuver2A1,label=center:{\scriptsize $\bm{a}^0$}] {};
  \node[node_round, fill = black!0] (state2A1) [below left=0.1cm and 1.8cm of action2A1,label=center:{\scriptsize $\bm{x}^0$}] {};
  \node[node_round, fill = black!0] (measurement2A1) [above=0.4cm of state2A1,label=center:{\scriptsize $\bm{z}^0$}] {};
  \node[rectangle, inner sep=2.1mm,draw=black!40,fill=black!30, fill opacity=0.0, fit= (route2A1) (maneuver2A1) (action2A1) (state2A1)] {};

  \coordinate (state2A1_a) at ($(state2A1)+(0.3,0)$);
  \coordinate (state2A1_b) at ($(state2A1)-(0.3,0)$);
  \coordinate (route2A1_b) at ($(route2A1)-(0.3,0)$);
  \coordinate (maneuver2A1_b) at ($(maneuver2A1)-(0.3,0)$);
  \coordinate (action2A1_b) at ($(action2A1)-(0.3,0)$);

  \path (route2A1) edge [connect] (maneuver2A1);
  \path (maneuver2A1) edge [connect] (action2A1); 
  \path (route2A1) edge [bend left, connect] (action2A1);
  \path (state2A1) edge [connect] (route2A1);
  \path (state2A1) edge [connect] (maneuver2A1);
  \path (state2A1) edge [connect] (action2A1);
  \path (state2A1) edge [connect] (measurement2A1);

  \node[node_round,fill = black!0] (route2A2) [below=1.65cm of action2A1,label=center:{\scriptsize $r^K$}] {};
  \node[node_round,fill = black!0] (maneuver2A2) [below=0.3cm of route2A2,label=center:{\scriptsize $m^K$}] {};
  \node[node_round,fill = black!0] (action2A2) [below=0.3cm of maneuver2A2,label=center:{\scriptsize $\bm{a}^K$}] {};
  \node[node_round,fill = black!0] (state2A2) [below left=0.1cm and 1.8cm of action2A2,label=center:{\scriptsize $\bm{x}^K$}] {};
  \node[node_round, fill = black!0] (measurement2A2) [above=0.4cm of state2A2,label=center:{\scriptsize $\bm{z}^K$}] {};
  \node[rectangle, inner sep=2.1mm,draw=black!40,fill=black!30, fill opacity=0.0, fit= (route2A2) (maneuver2A2) (action2A2) (state2A2)] {};

  \coordinate (state2A2_a) at ($(state2A2)+(0.3,0)$);
  \coordinate (state2A2_b) at ($(state2A2)-(0.3,0)$);
  \coordinate (route2A2_b) at ($(route2A2)-(0.3,0)$);
  \coordinate (maneuver2A2_b) at ($(maneuver2A2)-(0.3,0)$);
  \coordinate (action2A2_b) at ($(action2A2)-(0.3,0)$);

  \path (route2A2) edge [connect] (maneuver2A2);
  \path (maneuver2A2) edge [connect] (action2A2);
  \path (route2A2) edge [bend left, connect] (action2A2);
  \path (state2A2) edge [connect] (route2A2);
  \path (state2A2) edge [connect] (maneuver2A2);
  \path (state2A2) edge [connect] (action2A2);
  \path (state2A2) edge [connect] (measurement2A2);

  \path (state2A1) edge [connect] (maneuver2A2);
  \path (state2A1) edge [connect] (action2A2);
  
  \path (state2A2) edge [connect] (maneuver2A1);
  \path (state2A2) edge [connect] (action2A1);

  \path (stateA1) edge [connect, densely dashed] (state2A1);
  \path (actionA1) edge [connect, densely dashed] (state2A1);
  \path (routeA1) edge [connect, densely dashed] (route2A1);
  \path (maneuverA1) edge [connect, densely dashed] (maneuver2A1);
  
    \path (stateA2) edge [connect, densely dashed] (state2A2);
  \path (actionA2) edge [connect, densely dashed] (state2A2);
  \path (routeA2) edge [connect, densely dashed] (route2A2);
  \path (maneuverA2) edge [connect, densely dashed] (maneuver2A2);
  
  \node (dots1) [rotate=90, below left=0.27cm and -0.03cm of stateA1] {\scriptsize $\bm{\cdots}$};
  \node (dots2) [rotate=90, below left=0.27cm and -2.27cm of stateA1] {\scriptsize $\bm{\cdots}$};
  \node (dots3) [rotate=90, below left=0.27cm and -4.64cm of stateA1] {\scriptsize $\bm{\cdots}$};
  \node (dots4) [rotate=90, below left=0.27cm and -6.9cm of stateA1] {\scriptsize $\bm{\cdots}$};

  \node[rectangle, thick, draw =black!0, fill = black!10, minimum width=1.5cm, minimum height = 0.2cm] (time1) [above left =0.43cm and 0.12cm  of routeA1] {\tiny time step $t$ };  
  \node[rectangle, thick, draw =black!0, fill = black!10, minimum width=1.5cm, minimum height = 0.2cm] (time1) [above left =0.43cm and 0.12cm  of route2A1] {\tiny time step $t+1$};

\end{tikzpicture}}
\caption{\gls{DBN} showing the interdependencies between agents.
Random variables are drawn as circles, causal and temporal dependencies as solid and dashed arrows, respectively. $r$, $m$, and $\bm{a}$ also depend on the map.
}
\label{fig:dbn}
\vspace{-1px}
\end{figure}

In this work, we model the development of a traffic scene as a Markov process in the form of a \gls{DBN}, consisting of all agents in a scene.
This allows to explicitly model relations between agents, include domain knowledge and handle the uncertainty of measurements and human behavior.
Each agent follows its own decision making process, which is divided into three hierarchical layers: the route intention, the maneuver intention and the continuous action.
The random variables of the presented \gls{DBN} and their causal and temporal dependencies are depicted in \Fref{fig:dbn} and are explained in detail later in this section. In order to account for changing situations, the network structure is adapted online (creating and deleting agents as well as route and maneuver hypotheses). Thus, it can be applied to varying situations with an arbitrary number of agents, intention hypotheses and different road layouts.
As our DBN describes a hybrid, non-linear system with a multi-modal, non-Gaussian belief, sequential importance resampling is used for inference, allowing to represent arbitrary probability distributions.

\subsection{Estimation and Prediction}

\setlength{\timeslicedist}{1.3cm}
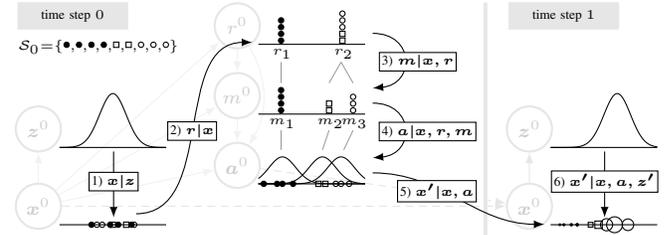
\begin{figure}
      \vspace{5px}
\centering
\pgfmathdeclarefunction{gauss}{2}{%
  \pgfmathparse{1/(#2*sqrt(2*pi))*exp(-((x-#1)^2)/(2*#2^2))}%
}
      \hspace{-8px}
\scalebox{1}{
\begin{tikzpicture}

\tikzstyle{samplesr1m1}=[fill, draw, circle, minimum width=2pt, inner sep=0pt]
\tikzstyle{samplesr2m1}=[draw, rotate=45, diamond, minimum width=3pt, minimum height=3pt, inner sep=0pt]
\tikzstyle{samplesr2m2}=[draw, circle, minimum width=2pt, inner sep=0pt]

\tikzstyle{node_round}=[circle, minimum size =6mm, thick, draw =black!10, node distance = 5mm]
\tikzstyle{node_square}=[rectangle, minimum size = 6mm, thick, draw =black!80, node distance = 5mm]
\tikzstyle{connect}=[-latex]
\tikzstyle{box}=[rectangle, draw=black!100]

  \node[node_round,fill = black!0] (routeA1) [label={[black!20]center:{\scriptsize $r^0$}}] {};
  \node[node_round,fill = black!0] (maneuverA1) [below=0.3cm of routeA1,label={[black!20]center:{\scriptsize $m^0$}}] {};
  \node[node_round,fill = black!0] (actionA1) [below=0.3cm of maneuverA1,label={[black!20]center:{\scriptsize $\bm{a}^0$}}] {};
  \node[node_round, fill = black!0] (stateA1) [below left=0.1cm and 2.2cm of actionA1,label={[black!20]center:{\scriptsize $\bm{x}^0$}}] {};
  \node[node_round, fill = black!0] (measurementA1) [above=0.4cm of stateA1,label={[black!20]center:{\scriptsize $\bm{z}^0$}}] {};

  \coordinate (stateA1_a) at ($(stateA1)+(0.3,0)$);
  \coordinate (routeA1_b) at ($(routeA1)-(0.3,0)$);
  \coordinate (routeA1_a) at ($(routeA1)+(0.3,0)$);
  \coordinate (maneuverA1_b) at ($(maneuverA1)-(0.3,0)$);
  \coordinate (maneuverA1_a) at ($(maneuverA1)+(0.3,0)$);
  \coordinate (actionA1_b) at ($(actionA1)-(0.3,0)$);
  \coordinate (actionA1_a) at ($(actionA1)+(0.3,0)$);

  \path (routeA1) edge [connect, black!5] (maneuverA1);
  \path (maneuverA1) edge [connect, black!5] (actionA1); 
  \path (routeA1) edge [bend left, connect, black!5] (actionA1);
  \path (stateA1) edge [connect, black!5] (routeA1);
  \path (stateA1) edge [connect, black!5] (maneuverA1);
  \path (stateA1) edge [connect, black!5] (actionA1);
  \path (stateA1) edge [connect, black!5] (measurementA1);

  \node[node_round, fill = black!0] (state2A1) [right=5.9cm of stateA1,label={[black!20]center:{\scriptsize $\bm{x}^0$}}] {};
  \node[node_round, fill = black!0] (measurement2A1) [above=0.4cm of state2A1,label={[black!20]center:{\scriptsize $\bm{z}^0$}}] {};

  \coordinate (state2A1_a) at ($(state2A1)+(0.3,0)$);
  \coordinate (state2A1_b) at ($(state2A1)-(0.3,0)$);
  \coordinate (route2A1_b) at ($(route2A1)-(0.3,0)$);
  \coordinate (maneuver2A1_b) at ($(maneuver2A1)-(0.3,0)$);
  \coordinate (action2A1_b) at ($(action2A1)-(0.3,0)$);

  \path (state2A1) edge [connect, black!10] (measurement2A1);

  \path (stateA1) edge [connect, densely dashed, black!10] (state2A1);
  \path (actionA1) edge [connect, densely dashed, black!10] (state2A1);

  \draw[black!10, line width=0.5mm] (3.3cm,0.31) --(3.3cm,-2.8);
  \node[rectangle, thick, draw =black!0, fill = black!10, minimum width=1.5cm, minimum height = 0.2cm] (time1) [above left =-0.3cm and 1.2cm  of routeA1] {\tiny time step $0$ };  
  \node[rectangle, thick, draw =black!0, fill = black!10, minimum width=1.5cm, minimum height = 0.2cm] (time2) [right = 5.cm  of time1] {\tiny time step $1$};

\begin{axis}[
  at={($(measurementA1)+(8,-7)$)},
  domain=-2:2,
  axis lines*=middle,
  axis y line=none,
  xtick=\empty,
  ytick=\empty,
  xlabel style={right},
  height=2.3cm, width=3cm,
  clip=false,
  axis on top,
  samples=200,
  xmin=-2,
  xmax=2,
  ymin=0,
  ymax=0.8]
  \addplot [black] {gauss(.0,0.5)};
\end{axis}

\begin{axis}[
  at={($(stateA1)+(8,-7)$)},
  domain=-2:2,
  axis lines*=middle,
  axis y line=none,
  xtick=\empty,
  ytick=\empty,
  xlabel style={right},
  height=2.3cm, width=3cm,
  clip=false,
  axis on top,
  samples=200,
  xmin=-2,
  xmax=2,
  ymin=0,
  ymax=0.8]
  \node [samplesr1m1] at (-0.8,0) {};
  \node [samplesr1m1] at (-0.1,0) {};
  \node [samplesr1m1] at (0.2,0) {};
  \node [samplesr1m1] at (0.7,0) {};
  \node [samplesr2m1] at (0.5,0) {};
  \node [samplesr2m1] at (0.,0) {};
  \node [samplesr2m2] at (-0.4,0) {};
  \node [samplesr2m2] at (-0.6,0) {};
  \node [samplesr2m2] at (0.8,0) {};
\end{axis}

\begin{axis}[
  at={($(routeA1)+(8,-7)$)},
  domain=-2:2,
  axis lines*=middle,
  axis y line=none,
  xtick=\empty,
  ytick=\empty,
  xlabel style={right},
  height=2.3cm, width=3cm,
  clip=false,
  axis on top,
  samples=200,
  xmin=-2,
  xmax=2,
  ymin=0,
  ymax=0.8]
  \node [] at (-1.1,-0.14) {\tiny $r_1$};
  \node [] at (1.2,-0.14) {\tiny $r_2$};
  \node [samplesr1m1] at (-1.15,0.055) {};
  \node [samplesr1m1] at (-1.15,0.155) {};
  \node [samplesr1m1] at (-1.15,0.255) {};
  \node [samplesr1m1] at (-1.15,0.355) {};
  \node [samplesr2m1] at (1.15,0.055) {};
  \node [samplesr2m1] at (1.15,0.155) {};
  \node [samplesr2m2] at (1.15,0.255) {};
  \node [samplesr2m2] at (1.15,0.355) {};
  \node [samplesr2m2] at (1.15,0.455) {};
\end{axis}

\begin{axis}[
  at={($(maneuverA1)+(8,-7)$)},
  domain=-2:2,
  axis lines*=middle,
  axis y line=none,
  xtick=\empty,
  ytick=\empty,
  xlabel style={right},
  height=2.3cm, width=3cm,
  clip=false,
  axis on top,
  samples=200,
  xmin=-2,
  xmax=2,
  ymin=0,
  ymax=0.8]
  \draw[line width=0.1mm, black!40](-1.15,0.77)--(-1.15,0.46);
  \draw[line width=0.1mm, black!40](1.1,0.77)--(0.65,0.46);
  \draw[line width=0.1mm, black!40](1.1,0.77)--(1.55,0.46);
  \draw[line width=0.1mm, black!40](-1.15,-0.24)--(-1.15,-0.53);
  \draw[line width=0.1mm, black!40](0.65,-0.24)--(0.4,-0.53);
  \draw[line width=0.1mm, black!40](1.55,-0.24)--(1.2,-0.53);
  \node [] at (-1.1,-0.14) {\tiny $m_1$};
  \node [] at (0.7,-0.14) {\tiny $m_2$};
  \node [] at (1.6,-0.14) {\tiny $m_3$};
  \node [samplesr1m1] at (-1.15,0.055) {};
  \node [samplesr1m1] at (-1.15,0.155) {};
  \node [samplesr1m1] at (-1.15,0.255) {};
  \node [samplesr1m1] at (-1.15,0.355) {};
  \node [samplesr2m1] at (0.65,0.055) {};
  \node [samplesr2m1] at (0.65,0.155) {};
  \node [samplesr2m2] at (1.55,0.055) {};
  \node [samplesr2m2] at (1.55,0.155) {};
  \node [samplesr2m2] at (1.55,0.255) {};
\end{axis}

\begin{axis}[
  at={($(actionA1)+(8,-7)$)},
  domain=-2:2,
  axis lines*=middle,
  axis y line=none,
  xtick=\empty,
  ytick=\empty,
  xlabel style={right},
  height=2.3cm, width=3cm,
  clip=false,
  axis on top,
  samples=200,
  xmin=-2,
  xmax=2,
  ymin=0,
  ymax=1.6]
  \addplot [black!100] {gauss(-1.1,0.5)};
  \addplot [black!100] {gauss(0.4,0.5)};
  \addplot [black!100] {gauss(1.1,0.5)};
  \node [samplesr1m1] at (-1.8,0) {};
  \node [samplesr1m1] at (-1.3,0) {};
  \node [samplesr1m1] at (-1.1,0) {};
  \node [samplesr1m1] at (-0.7,0) {};
  \node [samplesr2m1] at (0.25,0) {};
  \node [samplesr2m1] at (0.51,0) {};
  \node [samplesr2m2] at (0.9,0) {};
  \node [samplesr2m2] at (1.1,0) {};
  \node [samplesr2m2] at (1.4,0) {};
\end{axis}

\begin{axis}[
  at={($(measurement2A1)+(8,-7)$)},
  domain=0:4,
  axis lines*=middle,
  axis y line=none,
  xtick=\empty,
  ytick=\empty,
  xlabel style={right},
  height=2.3cm, width=3cm,
  clip=false,
  axis on top,
  samples=200,
  xmin=0,
  xmax=4,
  ymin=0,
  ymax=0.8]
  \addplot [black] {gauss(2.5,0.5)};
\end{axis}

\begin{axis}[
  at={($(state2A1)+(8,-7)$)},
  domain=0:4,
  axis lines*=middle,
  axis y line=none,
  xtick=\empty,
  ytick=\empty,
  xlabel style={right},
  height=2.3cm, width=3cm,
  clip=false,
  axis on top,
  samples=200,
  xmin=0,
  xmax=4,
  ymin=0,
  ymax=0.8]
  \node [samplesr1m1, scale=0.35] at (0.35,0) {};
  \node [samplesr1m1, scale=0.4] at (0.5,0) {};
  \node [samplesr1m1, scale=0.45] at (0.78,0) {};
  \node [samplesr1m1, scale=0.5] at (1.0,0) {};
  \node [samplesr2m1, scale=0.9] at (1.5,0) {};
  \node [samplesr2m1, scale=1.4] at (1.8,0) {};
  \node [samplesr2m2, scale=2.3] at (2.1,0) {};
  \node [samplesr2m2, scale=3] at (2.4,0) {};
  \node [samplesr2m2, scale=2.2] at (2.9,0) {};
\end{axis}

  \path ($(measurementA1)+(1,-0.3)$) edge [connect, black] ($(stateA1)+(1,-0.15)$);
  \node [inner sep=1pt, draw, fill=white] at ($(measurementA1)+(1.0,-0.7)$) {\tiny 1) $\bm{x|z}$};
    
  \path ($(stateA1)+(1.3,-0.1)$) edge [connect, black, out=0,in=180,looseness=1] ($(routeA1)+(0.2,-0.22)$);
  \node [inner sep=1pt, draw, fill=white] at ($(routeA1)+(-0.6,-1.4)$) {\tiny 2) $\bm{r|x}$};  
  
  \path ($(routeA1)+(1.8,-0.1)$) edge [connect, black, , out=0,in=5,looseness=2] ($(maneuverA1)+(1.8,+0.1)$);
  \node [inner sep=1pt, draw, fill=white] at ($(maneuverA1)+(2.4,0.45)$) {\tiny 3) $\bm{m|x,r}$};

  \path ($(maneuverA1)+(1.8,-0.1)$) edge [connect, black, out=0,in=5,looseness=2] ($(actionA1)+(1.8,+0.1)$);
  \node [inner sep=1pt, draw, fill=white] at ($(actionA1)+(2.53,0.45)$) {\tiny 4) $\bm{a|x,r,m}$};

  \path ($(actionA1)+(1.8,-0.1)$) edge [connect, black, out=0,in=180,looseness=1] ($(state2A1)+(0.2,-0.24)$);  
  \node [inner sep=1pt, draw, fill=white] at ($(state2A1)+(-1.23,0.2)$) {\tiny 5) $\bm{x'|x,a}$};
  
  \path ($(measurement2A1)+(1,-0.3)$) edge [connect, black] ($(state2A1)+(1,-0.15)$);  
  \node [inner sep=1pt, draw, fill=white] at ($(measurement2A1)+(1.0,-0.7)$) {\tiny 6) $\bm{x'|x,a,z'}$};  

\newcommand{\markersamples}{\raisebox{0.5pt}{\tikz{\node[samplesr1m1](){};}}}
\newcommand{\markersampless}{\raisebox{0.5pt}{\tikz{\node[samplesr2m1](){};}}}
\newcommand{\markersamplesss}{\raisebox{0.5pt}{\tikz{\node[samplesr2m2](){};}}}
  \node [inner sep=1pt, fill=white] at ($(measurementA1)+(0.8,1.1)$) 
  {\tiny $\mathcal{S}_0 {=} \{\markersamples{,}\markersamples{,}\markersamples{,}\markersamples{,}\markersampless{,}\markersampless{,}\markersamplesss{,}\markersamplesss{,}\markersamplesss\}$};

\end{tikzpicture}}
\caption{Exemplary initial sample generation (1-4), motion prediction (5), and particle weighting (6), shown for a single agent. Distributions are depicted simplified as being one dimensional. 
One particle represents the complete state space, i.e., kinematic state, route, maneuver, and action.}
\label{fig:dbn_distributions}
\vspace{-9px}
\end{figure}

The goal of our framework is to estimate all drivers' intentions (route and maneuver) and to predict their future trajectories.
The general procedure is exemplarily depicted in \Fref{fig:dbn_distributions}:
Initially, a set of particles ${\mathcal{S}_0=\{S_0^1, \cdots,S_0^N\}}$, with ${S^i=[X^i,R^i,M^i,A^i]^\top}$ representing the complete scene, is sampled (steps 1-4) according to the measurement and the map: ${S_0^i \sim P(X_0,R_0,M_0,A_0|Z_0,\mathrm{map})}$, with corresponding weights ${\omega^i_0 = 1/N}$.
Then, each particle is predicted to the subsequent time step (step 5) according to the transition probability: ${S^i_{t+1} \sim P(S_{t+1}|S^i_t)}$.
As soon as a new measurement is available, the particle weights get updated according to the measurement likelihood (step 6): ${\omega^i_t = P(Z_t|X^i_t)~\omega^i_{t-1}}$.
The actual probability of a set of intentions $(R_t,M_t)$ is given by 
\begin{align}
P(R_t,M_t) = \frac{\sum_{j\in \mathcal{J}} \omega^j_t}{\sum_{i=1}^N \omega^i_t} \text{, } \mathcal{J} {:=} \{j~|~R_t^j{=}R_t, M_t^j{=}M_t\}.
\end{align}
For the intention estimation process, the \gls{DBN} is thus applied as a filter, comparing the different model hypotheses to the actual observations. The intention of a single agent can be derived through marginalization of the belief.

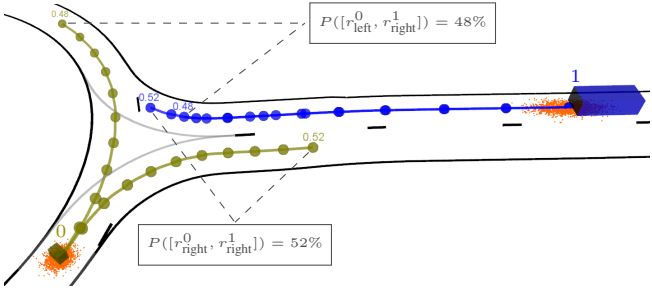
\begin{figure}
      \vspace{7px}
\centering
\begin{tikzpicture}
    \node[anchor=south west,inner sep=0] at (0,0) {\includegraphics[scale=0.17, trim=260 120 80 260, clip=true]{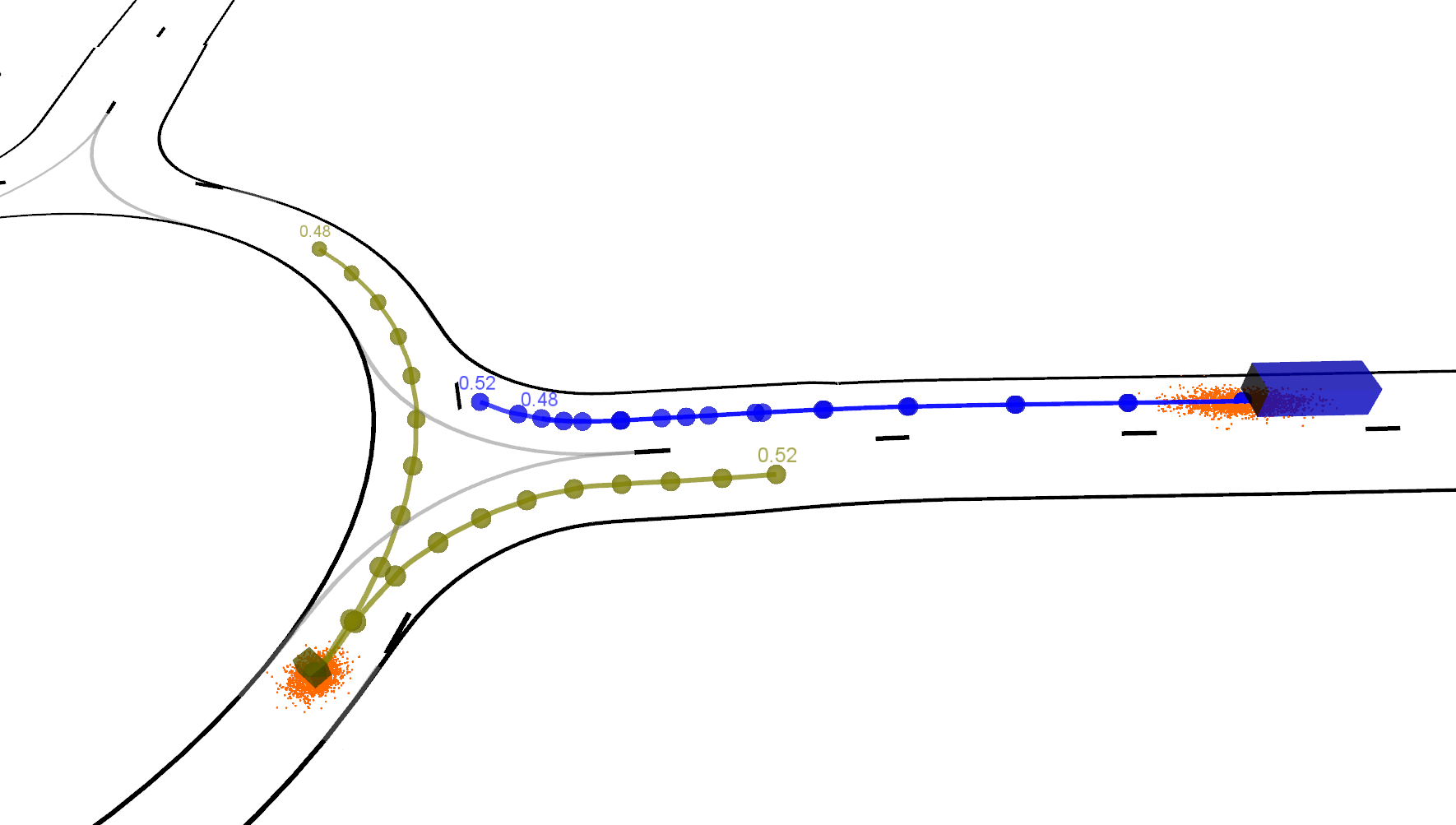}};
    \node[blue] at (7.6,2.8) {\scriptsize \textcolor{blue}{$1$}};
    \node[blue] at (0.76,0.73) {\scriptsize \textcolor{olive}{$0$}};
    \draw[black!70,dashed] (0.85,3.5) --(4,3.5);
    \draw[black!70,dashed] (2.46,2.31) --(4,3.5);
    \node[draw,black!70] at (5.3,3.5) {\tiny $P([r^{0}_\textrm{left},r^1_\textrm{right}])=48\%$};
    \draw[black!70,dashed] (1.98,2.3) --(3.07,0.85);
    \draw[black!70,dashed] (4.05,1.78) --(3.07,0.85);
    \node[draw,black!70] at (3.07,0.55) {\tiny $P([r^{0}_\textrm{right},r^1_\textrm{right}])=52\%$};
\end{tikzpicture}
\caption{Possible combinations of routes $[r^{0}_\textrm{left},r^1_\textrm{right}]$ and $[r^{0}_\textrm{right},r^1_\textrm{right}]$ at a roundabout: Although $\agent^1$ is currently not influenced by $\agent^{0}$, it has to slow down in the future if $\agent^{0}$ stays inside the roundabout. This influence has to be taken into account for the trajectory prediction using forward simulation.
}
\label{fig:routes_combinations}
\vspace{-9px}
\end{figure}

As DBNs are generative models, i.e., they can generate values of any of their random variables, it is possible to do a probabilistic forward simulation by iteratively predicting the current belief (including the estimated intentions) into subsequent time steps, applying the same models as for the filtering. 
As can be seen in \Fref{fig:dbn}, the action of an agent is modeled to be not \emph{directly} dependent on the actions or intentions of others, but only on its own intentions and the current context (given by $(\mathrm{map},X)$).
Thus, cyclic dependencies are avoided and one prediction step of an agent can be executed independently of the prediction steps of other agents.
However, as the current context depends on other agents' past actions, an interdependency between their trajectories emerges \emph{over time}, as shown in \Fref{fig:routes_combinations}.

In order to reduce complexity and improve interpretability of the trajectory prediction, the forward simulation is not done for each particle, but for the mean kinematic state of all agents given their route and maneuver intentions. 
For each \emph{combination} $(R,M)$ within $\mathcal{S}$, one multi-agent trajectory is generated and weighted with the corresponding probability $P(R,M)$.
Due to the interdependencies of multiple agents' future trajectories, this combinatorial aspect cannot be neglected within the prediction of the scene development.

\mytodo{
disadvantages:
particle degeneration
particle depletion
particle impoverishment
complexity}

\mytodo{
sampling according to utility functions (cumulative PDF and uniform sampling)
}

\mytodo{\begin{itemize}
\item formal definition of DBN (state space, belief space, actions, state, etc.)
\item online filtering (instead of unrolling)
\item Advantages of DBN modeling
\item Figure with nodes of DBN
\item Environment Model 
\item Subsections explain the single nodes
\end{itemize}}

\mytodo{Example for environment model \cite{lefevre_exploiting_2011} from road layout.}

The remainder of this section gives a detailed explanation of the single DBN nodes and their probability distributions.

\subsection{Vehicle Kinematics}
The action of each agent is defined as $\bm{a} = [a,\dot\theta]^\top$ with the longitudinal acceleration $a$ and the yaw rate $\dot{\theta}$.
It is the result of the decision making process, which is influenced by the current context and the agent's intentions, and is also estimated as a random variable of the \gls{DBN} (see \Fref{sec:sub_action_model}).
The transition of the kinematic state is given by the probability distribution $P(\bm{x}'|\bm{x},\bm{a}) = \mathcal{N}(\hat{\bm{x}'},\bm{Q})$, with
\begin{align}
\hat{\bm{x}'}=
\begin{pmatrix}
\hat{x'}\\
\hat{y'}\\
\hat{\theta'}\\
\hat{v'}\\
\end{pmatrix}
&{=}
\begin{pmatrix}
x + v \Delta T \cos(\theta') + \frac{1}{2} a \Delta T^2 \cos(\theta') \\
y + v \Delta T \sin(\theta') + \frac{1}{2} a \Delta T^2 \sin(\theta') \\
\theta + \dot{\theta} \Delta T \\
v + a \Delta T \\
\end{pmatrix}
\end{align}
and $\bm{Q} = \text{diag}(\sigma_x^2,\sigma_y^2,\sigma_\theta^2,\sigma_v^2)$. Although this model is simplistic, we argue that it is sufficient for prediction purposes.

\subsection{Measurement}
The proposed algorithm uses high-level cuboid objects as measurements, which can be derived by a magnitude of different sensors.
Hence, low-level sensor specifics are abstracted.
The data association, i.e., object detection and tracking, is handled by a separate algorithm and is considered to be given within this work.
The kinematic state $\bm{x}$ is measured with zero-mean Gaussian noise.
The measurement $\bm{z} = [x_z,y_z,\theta_z,v_z]^\top$ is distributed according to $P(\bm{z}|\bm{x}) = \mathcal{N}(\hat{\bm{z}},\bm{R})$, with $\hat{\bm{z}} = \bm{x}$ and ${\bm{R} = \text{diag}(\sigma_{z_x}^2,\sigma_{z_y}^2,\sigma_{z_\theta}^2,\sigma_{z_v}^2)}$.

\mytodo{check paper sources for particle filter measurement model, or thrun prob rob}
\mytodo{set evidence on measurement / state}

\subsection{Route Intention} \label{sec:sub_route_intention}

\begin{figure}
      \vspace{7px}
\centering
	\begin{picture}(300,70)
	\put(0,0){\includegraphics[scale=0.203,  trim=15 350 40 65, clip=true]{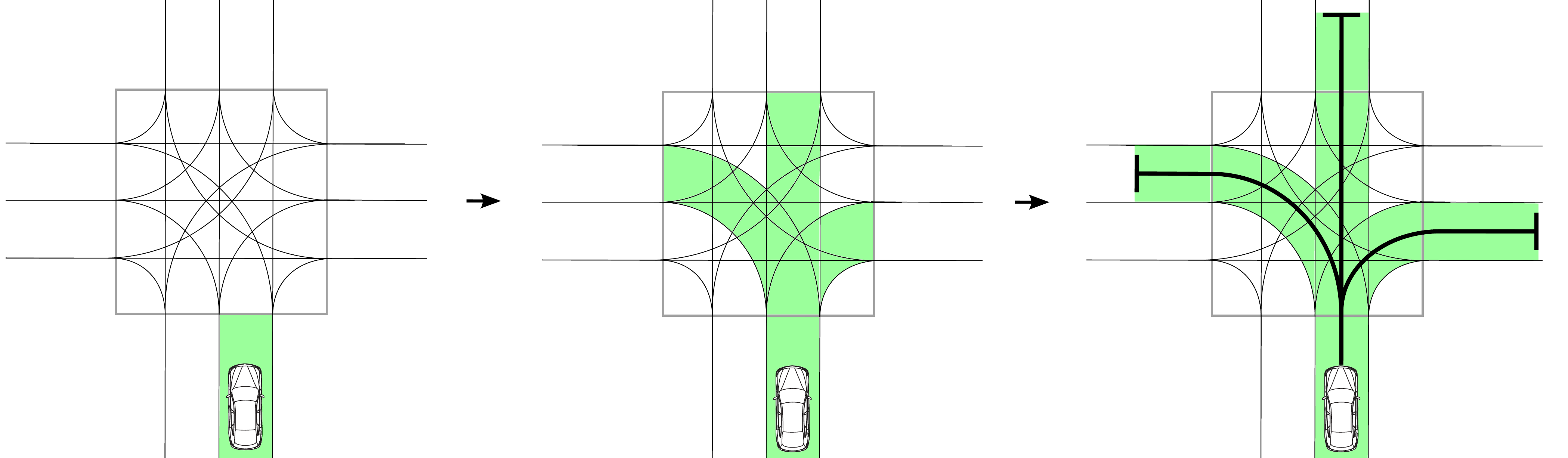}}
	\put(217,67){\tiny {$l_H$}}
	\put(175,53){\tiny {$l_H$}}
	\put(237,25){\tiny {$l_H$}}
	\end{picture}
\caption{Breadth-first search for possible routes of length $l_H$ on lane graph.
}
\label{fig:route_search}
\vspace{-9px}
\end{figure}
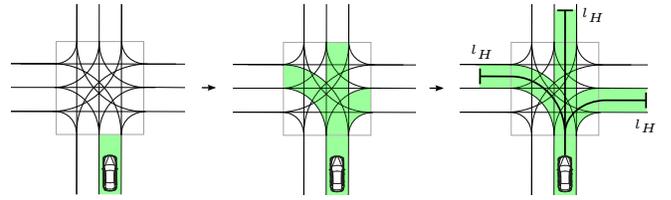

The route $r \in \mathcal{R}$ forms the first layer of an agent's decision making process and serves as a path that guides its behavior. It is represented by a sequence of consecutive lanes.
In every time step, the set of possible routes $\mathcal{R}$ is determined given the agent's pose, the topological map, and a specified metric horizon $l_H$. We apply breadth-first search on the lane graph starting with the current lane matching (see \Fref{fig:route_search}).

The route of an agent mainly serves two purposes:
Firstly, it allows to define relevant features along an agent's planned path such as the road curvature ahead or longitudinal distances to stop lines (see \Fref{sec:sub_action_model}).
Secondly, the routes of multiple agents allow to build relationships between agents on complex road layouts.
Two routes are related by dividing them into parts that either \emph{merge}, \emph{diverge}, \emph{cross}, are \emph{identical}, or have no relevant relation at all.
Different road junction types such as roundabouts, intersections or highway entrances can thus be broken down into these types of relations, allowing for a better generalization.
Typical relations between agents consist of distances to merging or crossing areas of their routes and corresponding right of way rules (see \Fref{sec:sub_action_model}).
As each route has a different geometry and may imply different traffic rules and relations to other agents, the route directly influences a driver's actions.

Initially, the desired route $r$ is sampled uniformly from the set of possible routes $\mathcal{R}$ according to ${P(r_i|\bm{x},\mathrm{map}) = |\mathcal{R}|^{-1},\quad \forall r_i \in \mathcal{R}}$.
Due to the fact that the route is only considered up to a specific horizon, a binary matching function ${s_r(r',r): \mathcal{R}' \times \mathcal{R} \longrightarrow \{0,1\}}$ is used to determine which of the routes $r' \in \mathcal{R}'$ are possible successors  of the current route $r$ (i.e., imply the same decisions at each contained intersection) and which are not.
If there are multiple candidates (in case of a route split), again, the route is sampled uniformly:
\begin{align}
P(r'_j|r_i,\bm{x},\mathrm{map}) = \frac{s_r(r'_j,r_i)}{\sum_{r' \in \mathcal{R'}}{s_r(r',r_i)}}.
\end{align}

\mytodo{
\begin{itemize}
\item how is it defined on intersection
\item how about lane changes, adjacent lanes?
\item within this paper: focus on intersecting lanes, not on lane change
\item why is route important: influences the short-term decision making
\item how is it represented: list of road sections
\end{itemize}}

\subsection{Maneuver Intention} \label{sec:sub_maneuver_intention}
\mytodo{
++++++++++++++++++++++++++++++++++++++++++++++++
FORMALIZE HOW WE DEFINE CONFLICT AREA! ALSO: Binary relation between two vehicles (pairwise)}

\begin{figure}
      \vspace{5px}
\centering
\begin{subfigure}{0.23\textwidth} 
\includegraphics[scale=0.34,  trim=420 350 530 120, clip=true]{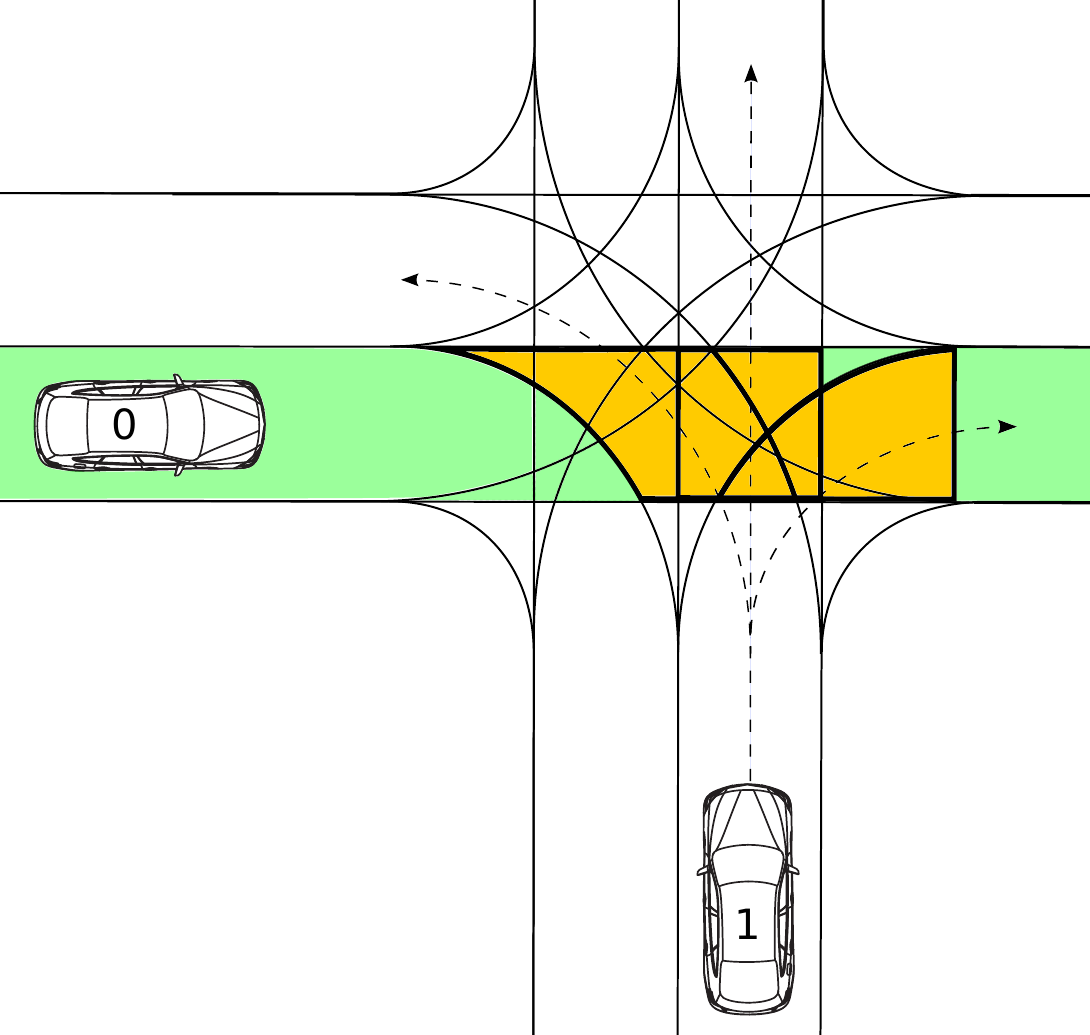}
\put(-105,5){\footnotesize (a)}
\end{subfigure}
\begin{subfigure}{0.245\textwidth} 
\centering
\includegraphics[scale=0.34,  trim=430 350 470 120, clip=true]{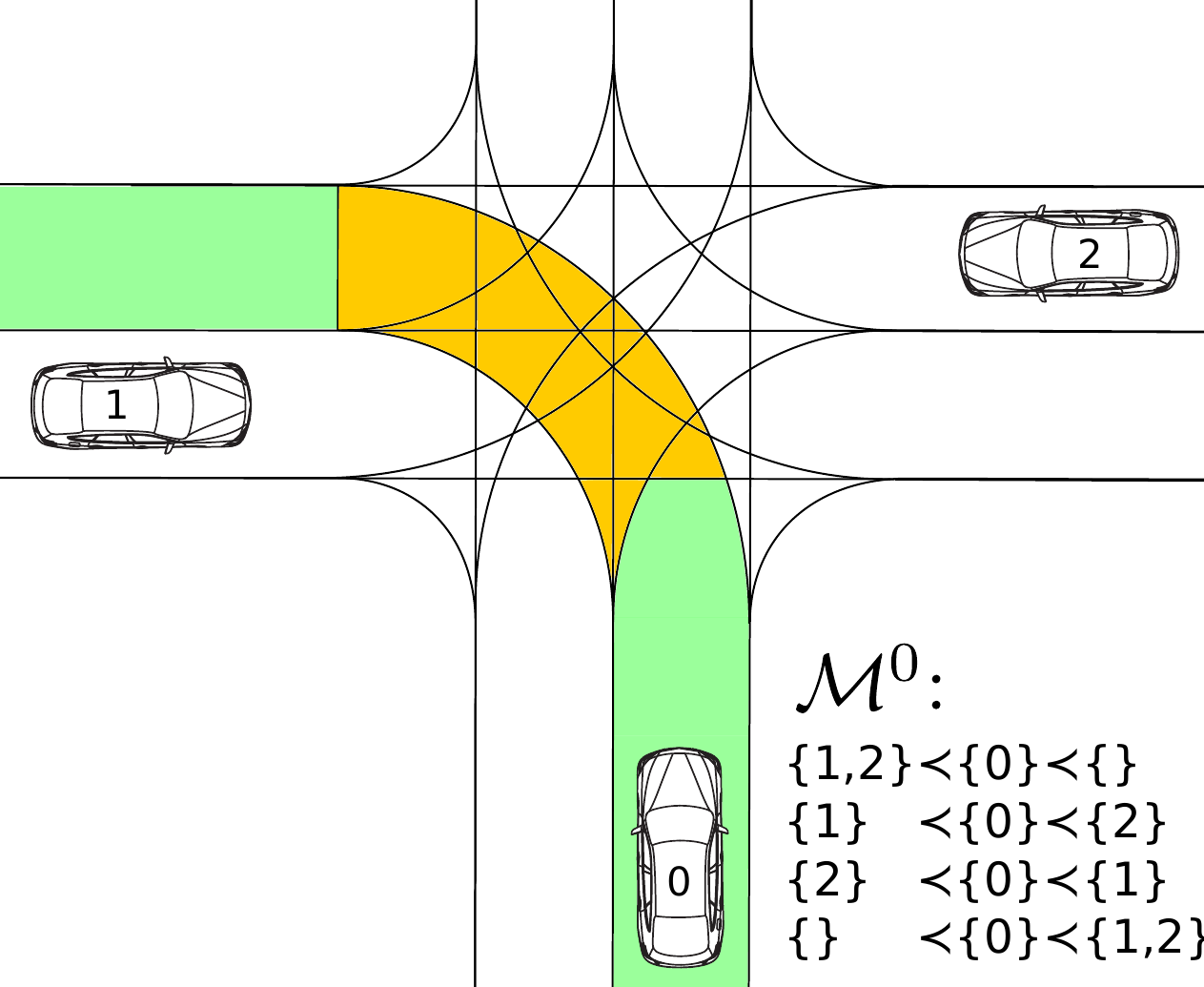}
\put(-122,5){\footnotesize (b)}
\end{subfigure}
\caption{(a): Possible conflict areas from $\agent^0$'s perspective for going straight, resulting from the three route hypotheses of $\agent^1$. The actual route of $\agent^1$ is unknown to $\agent^0$. (b): Four possible maneuvers for $\agent^0$ turning left, representing the sequence of agents passing the conflict areas.}
\vspace{-1px}
\label{fig:conflict_areas_and_maneuver_representation}
\end{figure}

The maneuver $m \in \mathcal{M}$ forms the second layer of the decision making process and describes the desired sequence, in which agents are going to merge or cross at intersections.
Therefore we introduce the notion of conflict areas:
\mytodo{mathematical definition of conflict area}
Given two agents on two routes, their conflict area is defined by the intersecting set of the areas of both routes, i.e., the area in which their lanes overlap.
We assume an agent doesn't know which route other agents are going to follow, thus, all possible conflict areas are considered (see \Fref{fig:conflict_areas_and_maneuver_representation}(a)). 

In order to avoid collisions at conflict areas, agents have to schedule their passing sequence.
A maneuver of agent $\agent^i$ states for all pairs $\langle \agent^i,\agent^j\rangle$ that have a \emph{potential} conflict (at least one route hypothesis of agent $\agent^j$ has a conflict with $\agent^i$'s intended route), whether $\agent^i$ will pass their conflict area first ($\agent^i {\prec} \agent^j$) or not ($\agent^i{\succ}\agent^j$).
This definition follows our previous work \cite{schulz_estimation_2017}, where maneuvers are based on the pseudo-homotopy of trajectories.
Our data suggests, that vehicles that have right of way are typically not influenced by other vehicles approaching the intersection. Thus, different maneuvers are only considered for vehicles that do not have right of way.
The set of possible maneuvers $\mathcal{M}$ can be derived given the agent's route, the map, and the kinematic states of all agents.
An example can be seen in \Fref{fig:conflict_areas_and_maneuver_representation}(b). 
A more detailed description of this concept of maneuvers, also including lane changes, can be found in \cite{schulz_estimation_2017}.
\mytodo{Unreasonable or impossible maneuvers are pruned using heuristics based on vehicle kinematics.} \mytodo{how}

The desired maneuver $m$ is initially sampled uniformly from the set of possible maneuvers $\mathcal{M}$ according to ${P(m_i|X,r,\mathrm{map}) = |\mathcal{M}|^{-1},\quad \forall m_i \in \mathcal{M}}$.
As situations change over time, the set of possible maneuvers may change as well (e.g., a new agent arrives or an existing agent traverses a conflict area). Hence, for further time steps, a matching function ${s_m(m',m): \mathcal{M}' \times \mathcal{M} \longrightarrow \{0,1\}}$ determines which of the new maneuvers $m' \in \mathcal{M}'$ are possible successors of the current maneuver $m$ (i.e., there are no contradictory passing sequences).
If there are multiple matching candidates, again, the maneuver is sampled uniformly.
\mytodo{explain better}

\mytodo{
\begin{itemize}
\item why is maneuver important: influences the short-term decision making AND longer prediction only possible for known maneuver (other-wise only physical prediction possible)
\item see survey \cite{lefevre_survey_2014} for distinction between model-based, physic-based and maneuver-based prediction
\item splits multimodal action distribution into (hopefully unimodal) parts
\item how is it represented: conflicting vehicles are separated in two lists: pre and after passing
\item how is it derived: reasonable acceleration range for own vehicle, constant velocity assumption for others, time gap
\end{itemize}}

\subsection{Action Model}\label{sec:sub_action_model}

\begin{table}
\vspace{5px}
\begin{center}
\setlength{\tabcolsep}{.68em}
\caption{Influences, Features, and Action Ranges for agent $\agent^i$}
\label{tab:context}
\begin{tabular}{llll}
\toprule
Influence && Features & Action Range\\
\midrule
vehicle dynamics && - & $[a_\mathrm{vd}^\mathrm{min},\> a_\mathrm{vd}^\mathrm{max}]$\\[0.3ex]
speed limit &&  $d_{v_\mathrm{lim}}$, $v_\mathrm{lim}$, $v^i$ & $[-\infty,\> a_\mathrm{IDM}^\mathrm{max}]$\\[0.3ex]
preceding agent $\agent^p$ && $d^p$, $v^p$, $v^i$ & $[-\infty, \> a_\mathrm{IDM}^\mathrm{max}]$\\[0.3ex]
road curvature && $d_\rho$, $\rho$, $v^i$ & $[-\infty,\> a_\mathrm{curve}^\mathrm{max}]$\\[0.3ex]
conflicting agent $\agent^c$ && $\chi^{i,c}$, $d_{\mathrm{entry}}^c$, $d_{\mathrm{exit}}^c$, $v^c$, & $[a_\mathrm{conf}^\mathrm{min},\> a_\mathrm{conf}^\mathrm{max}]$\\
&& $d_{\mathrm{yield}}^i$, $d_{\mathrm{entry}}^i$, $d_{\mathrm{exit}}^i$, $v^i$ & \\
\bottomrule
\end{tabular}
\end{center}
      \vspace{-1px}
\end{table}

\mytodo{ego vehicle velocity is also used!}

The action $\bm{a}=[a,\dot{\theta}]$ of an agent depends on his route and maneuver intentions, the kinematic states of all agents, and the map. It forms the third layer of the decision making process. 
Within this section, a heuristics-based probabilistic action model $P(\bm{a}|r,m,X,\mathrm{map})$ is defined to show the potential of the general framework. 
In order to narrow down the large number of dependencies, we define a set of submodels, each handling one so-called \emph{influence}. Each influence consists of a subset of the available features and constrains the acceleration to a range $[a^\mathrm{min},a^\mathrm{max}]$ that is plausible (e.g., not leading to collisions or violations of traffic rules) given that specific influence. 
\Fref{tab:context} shows the influences considered within this work, their corresponding features and action ranges.
These influences represent the context on which an agent's actions are based on and are derived deterministically given the variables of the \gls{DBN}.

\mytodo{figure of route features??}

\mytodo{
\begin{itemize}
\item transformation to increase generalization
\item why is context important: influences the short-term decision making
\item how is it represented: features that directly influence the action
\item how is it derived: map data and states, deterministically
\item easily extendable by additional features
\item reduce environment model and other agents to maneuver-relevant parts (or action-relevant parts)
\item deterministic transformation of static environment model and agent states given route and maneuver
\item allows generalization
\end{itemize}}

\mytodo{
\begin{itemize}
\item vehicle kinematics (max speed, max acceleration, max yaw rate (depending on velocity?))
\item speed limit compliance
\item preceding vehicle car following (distance and velocity)
\item intersection approach (traffic light, stop lines, yield lines, right of way, possible conflicts)
\item conflicting vehicles (real conflicts) $\rightarrow$ overlapping areas
\end{itemize}
}

\mytodo{
\begin{itemize}
\item just a lookup
\item maneuver and route influence context
\item influences: road geometry (curvature), preceding vehicle, traffic light, intersection, conflicting vehicles
\item knowledge of agents: in what extend are the goals of others known (not at all, probability distribution of filter, full knowledge)
\end{itemize}}

The influence \emph{vehicle dynamics} restricts the range of possible accelerations to the constant range $[a_\mathrm{vd}^\mathrm{min},\> a_\mathrm{vd}^\mathrm{max}]$.
\emph{Speed limits} are defined by a set of pairs of speed limit $v_\mathrm{lim}$ and distance along the route $d_{v_\mathrm{lim}}$ where it becomes effective.
A \emph{preceding agent} $\agent^p$ is described by its relative distance $d^p$ and its velocity $v^p$.
For both of these influences, the so-called \gls{IDM} presented by \cite{treiber_Congested_2000} is employed, dictating a maximum reasonable acceleration
\begin{align}\label{eq:idm}
a_\mathrm{IDM}^\mathrm{max} = a_\mathrm{d} \!\left( \!1 {-} \!\left(\frac{v^i}{v_\mathrm{lim}}\right)^{\!\!\delta} \!{-} \!\left(\frac{d_\mathrm{d} \!+\! v^i T_d \!+\! \frac{v^i(v^i-v^p)}{2 \sqrt{\lvert a_\mathrm{d} b_\mathrm{d} \rvert}}}{d^p}\right)^{\!\!\!2}\right)\!\!.
\end{align}
The parameters minimum spacing $d_\mathrm{d}$, desired time headway $T_d$, comfortable acceleration $a_\mathrm{d}$, braking deceleration $b_\mathrm{d}$, and acceleration exponent $\delta$ have to be specified. Although not part of the evaluation, the influences \emph{red traffic light} and \emph{stop sign} are also handled using \eqref{eq:idm} by setting $v^p=0$ and $d^p$ to the corresponding distance.

As the \gls{IDM} was primarily designed for highway scenarios, the curvature of the road as well as merging or intersecting lanes are not considered. Thus, we define the following models allowing the prediction in urban scenarios:
The model for the influence \emph{curvature} is based on a desired maximum lateral acceleration $a_\mathrm{lat}^\mathrm{max}$ that implies a maximum velocity $v_\rho = \sqrt{\rho a_\mathrm{lat}^\mathrm{max}}$
at a given curve radius $\rho$. 
The maximum acceleration of $\agent^i$ for one time step $\Delta T$ to still be able to reach the velocity $v_\rho$ at the corresponding distance $d_\rho$ with the comfortable braking deceleration $b_d$ is
\begin{align}\label{eq:vel_in_dist}
&a^\mathrm{max}_{v_{\rho},d_\rho} = \tilde{a} = \notag \\
&\frac{-2v + \Delta T b_d + \sqrt{ 4 v \Delta T b_d + \Delta T^2 b_d^2 - 8 b_d d_\rho + 4 v_{\rho}^{2}}}{2 \Delta T},
\end{align}
which can be determined by the following equations:
\scalebox{1.0}{
\hspace{-8px}
\begin{tikzpicture}
\footnotesize
      \draw[->] (0,0) -- (3.8,0) node[right] {};
      \draw[->] (0,0) -- (0,1.3) node[above] {};
      \draw[scale=1,domain=0:0.5,smooth,variable=\x,black] plot ({\x},{\x+0.5});
      \draw[scale=1,domain=0.5:3.5,smooth,variable=\y,black] plot ({\y},{-\y/4+1+1/8});

      \node[] at (-0.15,0.5) {$v$};

      \draw[scale=1,domain=0.0:3.5,smooth,variable=\y,gray,densely dashed] plot ({\y},{0.24});      
      \node[] at (-0.15,0.24) {$v_\rho$};
      
      \draw[scale=1,domain=0.0:0.5,smooth,variable=\y,gray, densely dashed] plot ({\y},{1});      
      \node[] at (-0.15,1) {$v_1$};

      \draw[densely dashed,gray] (0.5,0) -- (0.5,1) {};
      \draw [decorate,decoration={brace,mirror},gray] (0,-0.01) -- (0.5,-0.01) {};
      \node[] at (0.25,-0.24) {$\Delta T$};

      \draw[densely dashed,gray] (3.5,0) -- (3.5,0.24) {};
      \draw [decorate,decoration={brace,mirror},gray] (0.5,-0.01) -- (3.5,-0.01) {};
      \node[] at (2.05,-0.24) {$\Delta T_2$};
      
      \node[rotate=+45] at (0.15,0.83) {$\tilde{a}$};
      \node[rotate=-10] at (2.05,0.758) {$b_d$};

      \node [fill, draw, circle, minimum width=2pt, inner sep=0pt, pin={[pin distance=0.2cm, pin edge=gray]5:$d_1$}] at (0.5,1) {};
      
      \node [fill, draw, circle, minimum width=2pt, inner sep=0pt, pin={[pin distance=0.2cm, pin edge=gray]90:$d_\rho$}] at (3.5,0.24) {};
                  
      \node[] at (6.3,0.5) {$\begin{aligned}
      v_1 &= v + \tilde{a} \Delta T \\
      v_\rho &= v_1 + b_d \Delta T_2\\
      d_1 &= v \Delta T + \tfrac{1}{2} \tilde{a} {\Delta T}^2 \\
      d_\rho &= d_1 + v_1 \Delta T_2 + \tfrac{1}{2} b_d {\Delta T_2}^2
      \end{aligned}$};

\end{tikzpicture}
}
The smallest allowed acceleration of all curvature distance pairs along the route is used.
This results in a foresighted curvature approach.

\mytodo{As occlusions are not the focus of this work, we assume occlusions to always be there.}

The \emph{conflict} model is based on conflict areas at overlapping lanes where vehicles have to coordinate a specific sequence of passing. 
A conflict of agent $\agent^i$ with another agent $\agent^c$ is described by the right of way $\chi^{i,c}$, their velocities and distances to entering and exiting the conflict area and their distances to potential yield lines $d_{\mathrm{entry}}$, $d_{\mathrm{exit}}$, and $d_\mathrm{yield}$, respectively.
If agent $\agent^i$ has right of way, results indicate that it is sufficient to assume that it is not influenced by the other agent ($[a^\mathrm{min}_\mathrm{conf},a^\mathrm{max}_\mathrm{conf}] = [-\infty,\infty]$). 
If agent $\agent^i$ has to yield, it acts according to its desired maneuver $m$. 
Each agent that is going to pass before $\agent^i$ introduces an upper bound of acceleration ($a^\mathrm{max}_\mathrm{conf}$), each agent that is going to pass after $\agent^i$ introduces a lower bound ($a^\mathrm{min}_\mathrm{conf}$).
These accelerations are determined such that a minimum time gap between the two passing vehicles at the overlapping areas is ensured, assuming others drive with constant velocity.

\mytodo{implement and explain conflicts model}

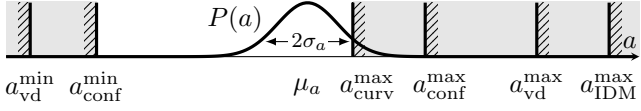
\begin{figure}
      \vspace{5px}
\pgfmathdeclarefunction{gauss}{2}{%
  \pgfmathparse{1/(#2*sqrt(2*pi))*exp(-((x-#1)^2)/(2*#2^2))}%
}
\hspace{-3px}
\begin{tikzpicture}
\def\myoffset{0.0}

\begin{axis}[
  domain=-1.2-2.3+\myoffset:10-3.08,
  axis x line=center,
  axis y line=none,
  xtick=\empty,
  ytick=\empty,
  xlabel style={right},
  every axis x label/.style={at={(current axis.right of origin)},above left=-0.1mm and -0.9mm},
  xlabel=$a$,
  height=2.3cm, width=10cm,
  clip=false,
  axis on top,
  samples=200,
  xmin=-1.2-2.3+\myoffset,
  xmax=10-3,
  ymin=0,
  ymax=0.8]

\fill[fill=black!0] (axis cs:1-3,0) rectangle (axis cs:6-3,0.82);
\fill[fill=black!10] (axis cs:-1.2-2.3+\myoffset,0) rectangle (axis cs:1-3+\myoffset,0.82);
\fill[fill=black!10] (axis cs:6-3-0.75,0) rectangle (axis cs:10-3,0.82);

\draw [yshift=+0.5cm](axis cs:3.3-3,0) node {$P(a)$};

  \addplot [very thick,black] {gauss(4.5-3,0.5)};
\draw [yshift=-0.4cm](axis cs:4.5-3,0) node {$\mu_a$};
\draw [yshift=+0.2cm, latex-latex](axis cs:3.8-3,0) -- node [] {} (axis cs:5.2-3,0);
\draw [yshift=+0.2cm, inner sep=0pt, fill=white](axis cs:4.5-3,0) node [fill=white] {\footnotesize $\,2\sigma_a$};

\draw[-,very thick,color=black] (axis cs:1-3+\myoffset,0) -- (axis cs:1-3+\myoffset,0.8);
\fill[pattern=north east lines, pattern color=black] (axis cs:1-3+\myoffset,0) rectangle (axis cs:0.8-3+\myoffset,0.8);
\draw [yshift=-0.4cm, latex-latex](axis cs:1-3+\myoffset,0) node [fill=white] {$a_{\mathrm{conf}}^\mathrm{min}$};

\draw[-,very thick,color=black] (axis cs:-0.1-3+\myoffset,0) -- (axis cs:-0.1-3+\myoffset,0.8);
\fill[pattern=north east lines, pattern color=black] (axis cs:-0.1-3+\myoffset,0) rectangle (axis cs:-0.3-3+\myoffset,0.8);
\draw [yshift=-0.4cm, latex-latex](axis cs:-0.1-3+\myoffset,0) node [fill=white] {$a_{\mathrm{vd}}^\mathrm{min}$};

\draw[-,very thick,color=black] (axis cs:6-3-0.75,0) -- (axis cs:6-3-0.75,0.8);
\fill[pattern=north east lines, pattern color=black] (axis cs:6-3-0.75,0) rectangle (axis cs:6.2-3-0.75,0.8);
\draw [yshift=-0.4cm, latex-latex](axis cs:6-3-0.5,0) node [fill=white] {$a_{\mathrm{curv}}^\mathrm{max}$};

\draw[-,very thick,color=black] (axis cs:7.2-3-0.75,0) -- (axis cs:7.2-3-0.75,0.8);
\fill[pattern=north east lines, pattern color=black] (axis cs:7.2-3-0.75,0) rectangle (axis cs:7.4-3-0.75,0.8);
\draw [yshift=-0.4cm, latex-latex](axis cs:7.2-3-0.5,0) node [fill=white] {$a_{\mathrm{conf}}^\mathrm{max}$};

\draw[-,very thick,color=black] (axis cs:8.3-3,0) -- (axis cs:8.3-3,0.8);
\fill[pattern=north east lines, pattern color=black ] (axis cs:8.3-3,0) rectangle (axis cs:8.5-3,0.8);
\draw [yshift=-0.4cm, latex-latex](axis cs:8.3-3,0) node [fill=white] {$a_{\mathrm{vd}}^\mathrm{max}$};

\draw[-,very thick,color=black] (axis cs:9.5-3,0) -- (axis cs:9.5-3,0.8);
\fill[pattern=north east lines, pattern color=black] (axis cs:9.5-3,0) rectangle (axis cs:9.7-3,0.8);
\draw [yshift=-0.4cm, latex-latex](axis cs:9.5-3,0) node [fill=white] {$a_{\mathrm{IDM}}^\mathrm{max}$};

\end{axis}

\end{tikzpicture}
\caption{Example of possible upper and lower bounds of the action models of the single influences, used to define the action probability distribution.
}
\label{fig:acc_range_combination}
\vspace{-9px}
\end{figure}

The ranges of feasible accelerations of the single influences are combined as shown in \Fref{fig:acc_range_combination} to the overall range
\begin{align}
a_\mathrm{max} &= \mathrm{min}\{a^\mathrm{max}_\mathrm{vd}, a^\mathrm{max}_\mathrm{curv}, a^\mathrm{max}_\mathrm{IDM}, a^\mathrm{max}_\mathrm{int},a^\mathrm{max}_\mathrm{conf}\},\\
a_\mathrm{min} &= \mathrm{max}\{a^\mathrm{min}_\mathrm{vd},a^\mathrm{min}_\mathrm{conf}\}.
\end{align}
Our measurement data suggests that drivers tend to minimize driving time while not exceeding the plausible acceleration range. Thus, accelerations are sampled from the distribution $P(a|r,m,X,\mathrm{map}) = \mathcal{N}(\mu_a,\sigma_a^2)$, with a mean close to the lowest maximum bound: $\mu_a = a_\mathrm{max} - {\sigma_a}$.
\mytodo{yaw rate, formula?}
The yaw rate is sampled from $P(\dot{\theta}|r,\bm{x},a,\mathrm{map}) = \mathcal{N}(\mu_{\dot{\theta}},\sigma_{\dot{\theta}}^2)$, given a mean yaw rate $\mu_{\dot{\theta}}$ that keeps the agent close to the center of its lane, which is calculated based on simple heuristics.
\mytodo{define k, g}

\mytodo{min max, ranges, utilities, distributions}

\mytodo{
\begin{itemize}
\item different influences from context form different utilities for action model
\item utilities allow violation of model, therefore more realistic
\item Intelligent Driver Model (for preceding vehicle, traffic light)
\item Gap maximization for conflicting
\item max lateral acceleration for curvatures
\item how to slow down before intersection?
\item sampling according to utility
\item interpretability and analyzability, one can inspect single features and their influences
\item figure with utilities
\item IDM specialities: don't allow negative velocity at full stops to prevent oscillation; also negative velocities are not possible as input as it might have a positive exponential (e.g. 4)
\item yaw rate within boundaries (depending on velocity?)
\item assume conservative model for multiple intersection conflicts (combine multiple possible conflict areas)
\item no fake knowledge is assumed!! distinguish between different models. (maybe future work)
\end{itemize}}

\mytodo{restrict a and yaw-rate to a specific range (dynamic constraints of vehicle?) and restrict a in a way that velocity is restricted (max a depends on current vel)}

\mytodo{restrict acceleration and yaw rate to a specific range (dynamic constraints of vehicle?)}

\section{Evaluation}

\newcommand{\mywidth}{1.0in}
\newcommand{\myheight}{0.5in}

\begin{figure*}
      \vspace{2px}
	\centering
	\footnotesize
	\begin{picture}(60,60)
	\put(0,10){\includegraphics[scale=0.15, trim=230 200 200 100, clip=true]{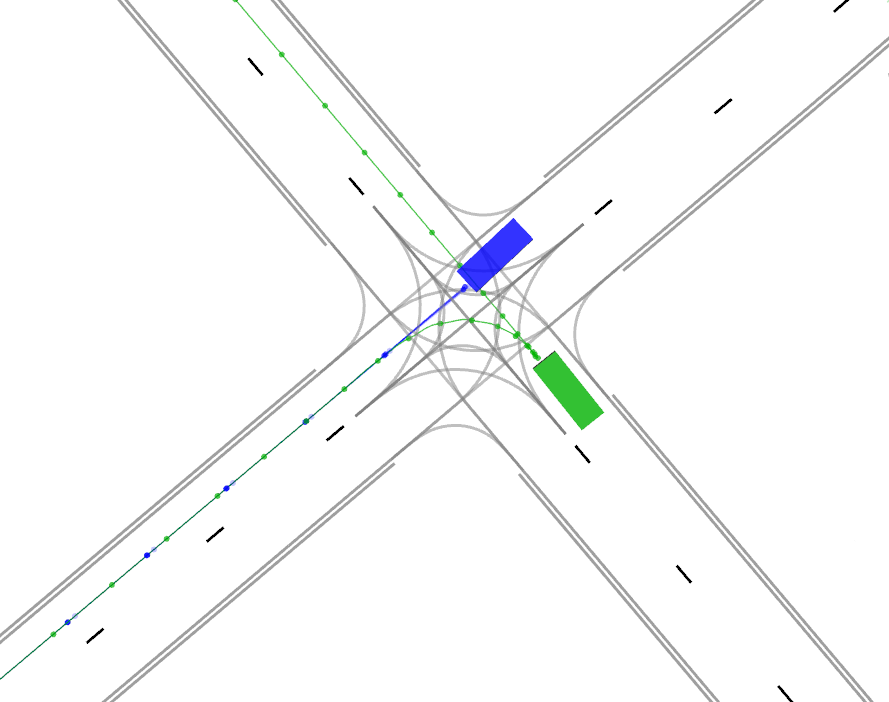}}
	\put(38,55){\scriptsize \textcolor{blue}{$1$}}
	\put(53,30){\scriptsize \textcolor{mycolor3}{$0$}}
	\end{picture}\hspace{0.2cm}
    \input{graphics/tikz/eval/new/scenario1again/m2_scene1.tikz}\hspace{0.1cm}
    \input{graphics/tikz/eval/new/scenario1again/m1_scene1.tikz}\hspace{0.1cm}
%
%
%
\begin{tikzpicture}

\begin{axis}[%
width=\mywidth,
height=\myheight,
at={(0.766in,0.486in)},
scale only axis,
xlabel={time [s]},
ylabel={$D_\mathrm{KL}$},
xmin=0,
xmax=22,
ymin=0,
ymax=1.8,
axis background/.style={fill=white},
title={route estimation error},
axis x line*=bottom,
axis y line*=left,
ytick={0.5, 1.5},
xtick={0, 10, 20},
legend style={legend cell align=left, align=left, draw=white!15!black}
]
\node[font=\sffamily,mycolor5] at (8.1,1.6) {\begin{tiny} {map-based} \end{tiny}};
\node[font=\sffamily,mycolor6] at (15,0.5) {\begin{tiny} {interactive} \end{tiny}};
\addplot [color=mycolor5]
  table[row sep=crcr]{%
0	1.08530127460838\\
0.21	1.08648610887027\\
0.42	1.07939806974431\\
0.62	1.07704651075355\\
0.82	1.07998682454494\\
1.02	1.07704651075355\\
1.22	1.0747004686218\\
1.43	1.06944191136836\\
1.64	1.06595150642859\\
1.85	1.05497825164808\\
2.06	1.04753901848414\\
2.26	1.03902349725648\\
2.46	1.03507426786415\\
2.66	1.03282454813013\\
2.86	1.02610559788139\\
3.06	1.01832345743933\\
3.26	1.01611106715639\\
3.46	1.01280163225503\\
3.66	1.0089544167376\\
3.86	1.00457564674175\\
4.06	1.00676263503826\\
4.26	1.00348394357656\\
4.4599999999999	1.00731013026135\\
4.6599999999999	1.00348394357656\\
4.8599999999999	1.00130410620114\\
5.0599999999999	0.998585974601012\\
5.2599999999999	1.0056685430254\\
5.4599999999999	1.0056685430254\\
5.6599999999999	1.00293853859985\\
5.8599999999999	0.998585974601012\\
6.0599999999999	1.00075988850063\\
6.2599999999999	1.00402964618219\\
6.4599999999999	1.00239343092759\\
6.6599999999999	0.999672340813233\\
6.8599999999999	0.998585974601012\\
7.0599999999999	0.996958634941637\\
7.2599999999999	0.998043233746441\\
7.4599999999999	0.999129010183187\\
7.6599999999999	0.999129010183187\\
7.8599999999999	0.997500787299725\\
8.0599999999999	1.00075988850063\\
8.2599999999999	1.00184862023583\\
8.4599999999999	1.0056685430254\\
8.6599999999999	1.00348394357656\\
8.8599999999999	1.00457564674175\\
9.0599999999999	1.00239343092759\\
9.2599999999998	0.999129010183187\\
9.4599999999998	0.992092445290464\\
9.6599999999998	0.984569572627476\\
9.8599999999998	0.959198232885458\\
10.06	0.940071260539584\\
10.26	0.853785526522781\\
10.46	0.765287858337671\\
10.66	0.710089901290298\\
10.86	0.694748461926959\\
11.06	0.698762919345621\\
11.26	0.731472294881655\\
11.46	0.81283202350597\\
11.66	0.891598119283808\\
11.86	1.01555873353285\\
12.06	1.14822278922144\\
12.26	1.33484107953098\\
12.46	1.55305757899799\\
12.66	1.68523962585038\\
12.86	1.73727128394399\\
13.06	1.46101790731583\\
13.26	1.06131650392444\\
13.46	0.773923490587276\\
13.66	0.699567748362908\\
13.86	0.659712404473747\\
14.06	0.642834366686413\\
14.26	0.648173814917252\\
14.46	0.689155159290448\\
14.66	0.767009029158444\\
14.86	0.944175935363716\\
15.06	1.1282470937978\\
15.26	1.31081589994398\\
15.46	1.47316029414155\\
15.66	1.53711725085447\\
15.86	1.46707067114718\\
16.06	1.3727860110951\\
16.26	1.25948551425619\\
16.46	1.15201306539522\\
16.66	1.11230561580011\\
16.86	1.09064411901893\\
17.06	1.07822159902038\\
17.26	1.10805674749611\\
17.46	1.06479074019263\\
17.66	1.05785429619597\\
17.86	0.977634362115234\\
18.06	0.885761526839356\\
18.26	0.820526109889821\\
18.46	0.72237003923689\\
18.66	0.657008133944111\\
18.86	0.636010989189193\\
19.06	0.588787165235757\\
19.26	0.561417409882103\\
19.46	0.555474375434925\\
19.66	0.470643834133223\\
19.86	0.338835025810472\\
20.06	0.207516295428584\\
20.26	0.114064956288051\\
20.46	0.0350056091988878\\
20.66	0.00501254182362483\\
20.86	8.01581023779331e-14\\
21.06	7.99360577730081e-14\\
21.260000000001	7.99360577730081e-14\\
21.460000000001	5.99520433297567e-14\\
21.660000000001	3.99680288865048e-14\\
21.860000000001	7.01660951563074e-14\\
22.060000000001	8.01581023779331e-14\\
22.260000000001	8.01581023779331e-14\\
22.460000000001	8.01581023779331e-14\\
22.660000000001	8.01581023779331e-14\\
};

\addplot [color=mycolor6]
  table[row sep=crcr]{%
0	1.07939806974425\\
0.21	1.08234531620424\\
0.42	1.08411784251766\\
0.62	1.07704651075355\\
0.82	1.07822159902038\\
1.02	1.07763388228292\\
1.22	1.07822159902038\\
1.43	1.07587280169862\\
1.64	1.07294454191956\\
1.85	1.06305211326959\\
2.06	1.05785429619597\\
2.26	1.05211045501672\\
2.46	1.04526111317114\\
2.66	1.04185395484953\\
2.86	1.04185395484953\\
3.06	1.03507426786415\\
3.26	1.03057987827631\\
3.46	1.02722229258146\\
3.66	1.02443289049389\\
3.86	1.02165124753201\\
4.06	1.02443289049389\\
4.26	1.01611106715639\\
4.4599999999999	1.01390356074118\\
4.6599999999999	1.00239343092759\\
4.8599999999999	0.99479296002914\\
5.0599999999999	0.986176859338348\\
5.2599999999999	0.986176859338348\\
5.4599999999999	0.982430534378727\\
5.6599999999999	0.978166135592269\\
5.8599999999999	0.972861083362576\\
6.0599999999999	0.971275040206941\\
6.2599999999999	0.974979728222861\\
6.4599999999999	0.968110480603978\\
6.6599999999999	0.972332122757806\\
6.8599999999999	0.968637212246391\\
7.0599999999999	0.970219073899737\\
7.2599999999999	0.968110480603978\\
7.4599999999999	0.97074691767024\\
7.6599999999999	0.966006324120114\\
7.8599999999999	0.960242619403364\\
8.0599999999999	0.955031560190612\\
8.2599999999999	0.950364712207932\\
8.4599999999999	0.9508821766438\\
8.6599999999999	0.944175935363716\\
8.8599999999999	0.937514368325807\\
9.0599999999999	0.934963997139802\\
9.2599999999998	0.933436890709151\\
9.4599999999998	0.930896884263543\\
9.6599999999998	0.933436890709151\\
9.8599999999998	0.934963997139802\\
10.06	0.972861083362576\\
10.26	0.968110480603978\\
10.46	0.945719542564993\\
10.66	0.907330990502708\\
10.86	0.878994947130482\\
11.06	0.876108942241347\\
11.26	0.866548639979365\\
11.46	0.807436326962118\\
11.66	0.795401414598746\\
11.86	0.77305656378829\\
12.06	0.750776293396624\\
12.26	0.719080562586491\\
12.46	0.702793557611791\\
12.66	0.693947500730754\\
12.86	0.689553645049855\\
13.06	0.696352311508934\\
13.26	0.699970405908111\\
13.46	0.714166547784062\\
13.66	0.722781985689677\\
13.86	0.728567125791247\\
14.06	0.730226188801408\\
14.26	0.731056753633207\\
14.46	0.732303895761181\\
14.66	0.73313618877635\\
14.86	0.73271995567981\\
15.06	0.728567125791247\\
15.26	0.729811164931578\\
15.46	0.735637232180939\\
15.66	0.737308076345757\\
15.86	0.734385928576584\\
16.06	0.737726223957955\\
16.26	0.73856304409047\\
16.46	0.73480285582882\\
16.66	0.732303895761181\\
16.86	0.718670137791072\\
17.06	0.728567125791247\\
17.26	0.714166547784062\\
17.46	0.726497160241541\\
17.66	0.7571525105359\\
17.86	0.768301854160749\\
18.06	0.737308076345757\\
18.26	0.755874010468293\\
18.46	0.758432647258617\\
18.66	0.788337367563738\\
18.86	0.817350784019297\\
19.06	0.795401414598746\\
19.26	0.718259881375061\\
19.46	0.702389761926918\\
19.66	0.60733652414689\\
19.86	0.438815087789781\\
20.06	0.275490031686485\\
20.26	0.154550759848668\\
20.46	0.0481403753280085\\
20.66	0.00944445882808053\\
20.86	0.0010005003336637\\
21.06	8.01581023779331e-14\\
21.260000000001	7.99360577730081e-14\\
21.460000000001	7.01660951563074e-14\\
21.660000000001	5.01820807130558e-14\\
21.860000000001	3.99680288865048e-14\\
22.060000000001	8.01581023779331e-14\\
22.260000000001	8.01581023779331e-14\\
22.460000000001	8.01581023779331e-14\\
22.660000000001	8.01581023779331e-14\\
};

\end{axis}
\end{tikzpicture}%
\hspace{0.1cm}
%
%
%
\begin{tikzpicture}

\begin{axis}[%
width=\mywidth,
height=\myheight,
at={(0.766in,0.486in)},
scale only axis,
xmin=0,
xmax=10,
ymin=0,
ymax=65,
axis background/.style={fill=white},
title style={xshift=-1.7ex,yshift=-0.59ex},
title={trajectory prediction error},
axis x line*=bottom,
axis y line*=left,
xlabel={prediction horizon [s]},
ylabel={$\epsilon^{(x,y)}$},
xtick={0, 5, 10},
ytick={0, 30, 60},
legend style={legend cell align=left, align=left, draw=white!15!black}
]
\node[rotate=23,font=\sffamily,mycolor4] at (6.4,24) {\begin{tiny} {CTRV} \end{tiny}};
\node[rotate=23,font=\sffamily,mycolor5] at (5,30) {\begin{tiny} {map-based} \end{tiny}};
\node[rotate=10,font=\sffamily,mycolor6] at (8.3,9) {\begin{tiny} {interactive} \end{tiny}};
\addplot [color=mycolor4]
  table[row sep=crcr]{%
0	2.46543205074455\\
0.5	3.80904278576748\\
1	5.40798492026462\\
1.5	6.96229065887844\\
2	8.60478650032282\\
2.5	10.0511315102167\\
3	11.9419514977117\\
3.5	13.8624104975605\\
4	16.1646413060544\\
4.5	18.5708763258988\\
5	21.2808187111483\\
5.5	24.1551524591471\\
6	27.2600237271052\\
6.5	30.6185127003207\\
7	34.1648561799946\\
7.5	38.0708417866671\\
8	42.0973314766765\\
8.5	46.5756861710521\\
9	51.0569940769504\\
9.5	55.9169910826623\\
10	60.7157196944598\\
};

\addplot [color=mycolor5]
  table[row sep=crcr]{%
0	1.81869969932856\\
0.5	2.81245849737755\\
1	4.07562694396598\\
1.5	5.62286657791431\\
2	7.50317260403849\\
2.5	9.72570155462905\\
3	12.1591531330899\\
3.5	14.6843751002698\\
4	17.6308518303905\\
4.5	20.6276373767119\\
5	23.8517174490415\\
5.5	26.7299353936409\\
6	30.0625895696401\\
6.5	33.2049774699119\\
7	36.8119919408644\\
7.5	40.059254949631\\
8	43.8702027092573\\
8.5	47.6270890044232\\
9	51.8268716864057\\
9.5	54.8911079613088\\
10	57.1642246709052\\
};

\addplot [color=mycolor6]
  table[row sep=crcr]{%
0	1.08486417838467\\
0.5	1.31725830877203\\
1	1.84792327406636\\
1.5	2.61346738879871\\
2	3.71048637791277\\
2.5	4.81019639029892\\
3	5.76124052992857\\
3.5	6.01858384825884\\
4	6.67209878481841\\
4.5	6.97365431188602\\
5	7.84281449218898\\
5.5	8.40914524301121\\
6	9.45147197149572\\
6.5	10.0546631116707\\
7	11.3177297163134\\
7.5	12.319637430908\\
8	13.9243810670746\\
8.5	15.4246618421734\\
9	17.3125720447741\\
9.5	19.0993833374957\\
10	21.4894465420722\\
};

\end{axis}
\end{tikzpicture}%
\\[2ex]
	\begin{picture}(60,60)
	\put(0,0){\includegraphics[scale=0.19, trim=100 70 10 50, clip=true]{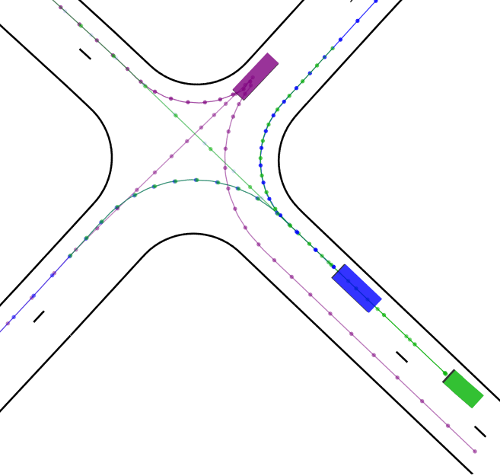}}
	\put(48,30){\scriptsize \textcolor{blue}{$1$}}
	\put(20,64){\scriptsize \textcolor{purple}{$2$}}
	\put(69,10){\scriptsize \textcolor{mycolor3}{$0$}}
	\end{picture}\hspace{0.2cm}
  \input{graphics/tikz/eval/new/scenario-4/route_prob_m2_scene-4_agent288.tikz}\hspace{0.1cm}
  \input{graphics/tikz/eval/new/scenario-4/route_prob_m1_scene-4_agent288.tikz}\hspace{0.1cm}
%
%
%
\begin{tikzpicture}

\begin{axis}[%
width=\mywidth,
height=\myheight,
at={(0.766in,0.486in)},
scale only axis,
xlabel={time [s]},
ylabel={$D_\mathrm{KL}$},
xmin=0,
xmax=21.6410000324249,
ymin=0,
axis background/.style={fill=white},
title style={font=\bfseries},
axis x line*=bottom,
axis y line*=left,
ytick={0.5, 1.5},
xtick={0, 10, 20},
legend style={legend cell align=left, align=left, draw=white!15!black}
]
\node[font=\sffamily,mycolor5] at (15,1.8) {\begin{tiny} {map-based} \end{tiny}};
\node[font=\sffamily,mycolor6] at (10,0.8) {\begin{tiny} {interactive} \end{tiny}};
\addplot [color=mycolor5]
  table[row sep=crcr]{%
0	1.09861228866811\\
0.200099945068359	1.0986122886682\\
0.410000085830688	1.0986122886682\\
0.640399932861328	1.0986122886682\\
0.920000076293945	1.0986122886682\\
1.11999988555908	1.0986122886682\\
1.3199999332428	1.0986122886682\\
1.54999995231628	1.0986122886682\\
1.75999999046326	1.0986122886682\\
2.00009989738464	1.0986122886682\\
2.20179986953735	1.0986122886682\\
2.4300000667572	1.0986122886682\\
2.64000010490417	1.0986122886682\\
2.83999991416931	1.0986122886682\\
3.04010009765625	1.0986122886682\\
3.24000000953674	1.0986122886682\\
3.50999999046326	1.0986122886682\\
3.71000003814697	1.0986122886682\\
3.91000008583069	1.0986122886682\\
4.11999988555908	1.0986122886682\\
4.3199999332428	1.0986122886682\\
4.54999995231628	1.0986122886682\\
4.75	1.0986122886682\\
4.95000004768372	1.0986122886682\\
5.15000009536743	1.0986122886682\\
5.36010003089905	1.0986122886682\\
5.58999991416931	1.0986122886682\\
5.78999996185303	1.0986122886682\\
5.99000000953674	1.0986122886682\\
6.19000005722046	1.0986122886682\\
6.39000010490417	1.0986122886682\\
6.59010004997253	1.0986122886682\\
6.78999996185303	1.0986122886682\\
6.99000000953674	1.0986122886682\\
7.19000005722046	1.09931253378259\\
7.40000009536743	1.08747454425771\\
7.62999987602234	1.09691373203259\\
7.82999992370605	1.09332628423893\\
8.02999997138977	1.08352665125013\\
8.23000001907349	1.08441356926836\\
8.4300000667572	1.11048246123866\\
8.62999987602234	1.17344359929118\\
8.82999992370605	1.26054315581437\\
9.03999996185303	1.37199705641911\\
9.24000000953674	1.47447005202366\\
9.51999998092651	1.57551969423069\\
9.75	1.62353683681365\\
9.95000004768372	1.61595912942442\\
10.1601998806	1.60793903631041\\
10.4000000953674	1.62709284767287\\
10.6299998760223	1.70210159729752\\
10.8299999237061	1.76200651107871\\
11.0299999713898	1.74354089695758\\
11.2300000190735	1.64454708954767\\
11.4400000572205	1.51594756934681\\
11.710000038147	1.40201732158517\\
11.9100000858307	1.3175148457251\\
12.1099998950958	1.23994546478482\\
12.3101000785828	1.19204423346073\\
12.5199999809265	1.17603343563676\\
12.75	1.17733087337958\\
12.9500000476837	1.1815592094891\\
13.1500000953674	1.18547822168984\\
13.3499999046326	1.18580550037055\\
13.5501000881195	1.18058185153543\\
13.75	1.16796236680297\\
13.9500000476837	1.15296288341825\\
14.1601998806	1.13974683202673\\
14.4017999172211	1.13134211329109\\
14.6400001049042	1.12300744550289\\
14.9100000858307	1.11779510808495\\
15.1617999076843	1.11443683901514\\
15.4000000953674	1.11382745447465\\
15.5999999046326	1.1150465951316\\
15.8399999141693	1.11657261404516\\
16.1099998950958	1.12331490130854\\
16.320100069046	1.13227249016081\\
16.5199999809265	1.14162417926146\\
16.7200999259949	1.15963699050595\\
16.9200999736786	1.17409042984413\\
17.1199998855591	1.1715056141885\\
17.3201999664307	1.14068506509008\\
17.5599999427795	1.07119169309218\\
17.7599999904633	0.930643238098632\\
17.960000038147	0.696552973694858\\
18.1603000164032	0.434327652793358\\
18.4000000953674	0.19322073766304\\
18.5999999046326	0.0570995920406681\\
18.8401000499725	0.00883894866729513\\
19.0401999950409	0.00130084573313819\\
19.2404999732971	0.000100005000423417\\
19.5204000473022	9.01501095995586e-14\\
19.7599999904633	9.01501095995586e-14\\
19.9600999355316	9.01501095995586e-14\\
20.1617999076843	8.90398865749336e-14\\
20.400899887085	6.99440505513824e-14\\
20.6400001049042	8.90398865749336e-14\\
20.8399999141693	9.01501095995586e-14\\
21.0401000976562	9.01501095995586e-14\\
21.2400000095367	9.01501095995586e-14\\
21.440299987793	9.01501095995586e-14\\
21.6410000324249	9.01501095995586e-14\\
};

\addplot [color=mycolor6]
  table[row sep=crcr]{%
0	1.09861228866811\\
0.200099945068359	1.0986122886682\\
0.410000085830688	1.0986122886682\\
0.640399932861328	1.0986122886682\\
0.920000076293945	1.0986122886682\\
1.11999988555908	1.0986122886682\\
1.3199999332428	1.0986122886682\\
1.54999995231628	1.0986122886682\\
1.75999999046326	1.0986122886682\\
2.00009989738464	1.0986122886682\\
2.20179986953735	1.0986122886682\\
2.4300000667572	1.0986122886682\\
2.64000010490417	1.0986122886682\\
2.83999991416931	1.0986122886682\\
3.04010009765625	1.0986122886682\\
3.24000000953674	1.0986122886682\\
3.50999999046326	1.0986122886682\\
3.71000003814697	1.0986122886682\\
3.91000008583069	1.0986122886682\\
4.11999988555908	1.0986122886682\\
4.3199999332428	1.0986122886682\\
4.54999995231628	1.0986122886682\\
4.75	1.0986122886682\\
4.95000004768372	1.0986122886682\\
5.15000009536743	1.0986122886682\\
5.36010003089905	1.0986122886682\\
5.58999991416931	1.0986122886682\\
5.78999996185303	1.0986122886682\\
5.99000000953674	1.0986122886682\\
6.19000005722046	1.0986122886682\\
6.39000010490417	1.0986122886682\\
6.59010004997253	1.0986122886682\\
6.78999996185303	1.0986122886682\\
6.99000000953674	1.09911241370988\\
7.19000005722046	1.10061429133887\\
7.40000009536743	1.08955344417982\\
7.62999987602234	1.0589109218166\\
7.82999992370605	1.02165124753204\\
8.02999997138977	1.01151758181723\\
8.23000001907349	1.00566854302543\\
8.4300000667572	1.00676263503829\\
8.62999987602234	1.00813193553979\\
8.82999992370605	1.01390356074121\\
9.03999996185303	1.02054075324801\\
9.24000000953674	1.02387594255415\\
9.51999998092651	1.01832345743936\\
9.75	1.03704693301001\\
9.95000004768372	1.06363131977292\\
10.1601998806	1.09541739777165\\
10.4000000953674	1.12732045280492\\
10.6299998760223	1.14633302487958\\
10.8299999237061	1.15677120284171\\
11.0299999713898	1.15327969001048\\
11.2300000190735	1.14319133206614\\
11.4400000572205	1.14381888150625\\
11.710000038147	1.14790758039305\\
11.9100000858307	1.16091331486555\\
12.1099998950958	1.17279718685755\\
12.3101000785828	1.18744350237479\\
12.5199999809265	1.19501306295453\\
12.75	1.21267724495615\\
12.9500000476837	1.2323722788477\\
13.1500000953674	1.24029106700012\\
13.3499999046326	1.24236719231332\\
13.5501000881195	1.23959998196916\\
13.75	1.23580752784602\\
13.9500000476837	1.21942490888493\\
14.1601998806	1.19567400151131\\
14.4017999172211	1.15423071243119\\
14.6400001049042	1.10473096967635\\
14.9100000858307	1.05153788128224\\
15.1617999076843	1.01362796476317\\
15.4000000953674	1.00402964618222\\
15.5999999046326	1.00731013026138\\
15.8399999141693	1.01998596821298\\
16.1099998950958	1.04611471971041\\
16.320100069046	1.08411784251772\\
16.5199999809265	1.12979340615175\\
16.7200999259949	1.18941265299467\\
16.9200999736786	1.24514208136408\\
17.1199998855591	1.26584820804406\\
17.3201999664307	1.25211317965394\\
17.5599999427795	1.18580550037055\\
17.7599999904633	1.01694014002414\\
17.960000038147	0.735637232180981\\
18.1603000164032	0.431706419313484\\
18.4000000953674	0.17304472001384\\
18.5999999046326	0.047196434831069\\
18.8401000499725	0.00682322534821607\\
19.0401999950409	0.00100050033367368\\
19.2404999732971	0.000200020002757155\\
19.5204000473022	9.01501095995586e-14\\
19.7599999904633	0.000100005000423417\\
19.9600999355316	0.000500125041772256\\
20.1617999076843	0.000700245114473604\\
20.400899887085	7.99360577730081e-14\\
20.6400001049042	0.000200020002757155\\
20.8399999141693	9.01501095995586e-14\\
21.0401000976562	9.01501095995586e-14\\
21.2400000095367	9.01501095995586e-14\\
21.440299987793	0.000100005000423417\\
21.6410000324249	9.01501095995586e-14\\
};

\end{axis}
\end{tikzpicture}%
\hspace{0.1cm}
%
%
\definecolor{mycolor1}{rgb}{0.00000,0.44700,0.74100}%
\definecolor{mycolor2}{rgb}{0.85000,0.32500,0.09800}%
\definecolor{mycolor3}{rgb}{0.92900,0.69400,0.12500}%
\begin{tikzpicture}

\begin{axis}[%
width=\mywidth,
height=\myheight,
at={(0.766in,0.486in)},
scale only axis,
xlabel={prediction horizon [s]},
ylabel={$\epsilon^{(x,y)}$},
xmin=0,
xmax=10,
ymin=0,
ymax=50,
axis background/.style={fill=white},
title style={font=\bfseries},
axis x line*=bottom,
axis y line*=left,
xtick={0, 5, 10},
legend style={legend cell align=left, align=left, draw=white!15!black}
]
\node[rotate=30,font=\sffamily,mycolor4] at (8.7,42) {\begin{tiny} {CTRV} \end{tiny}};
\node[rotate=18,font=\sffamily,mycolor5] at (8.8,28) {\begin{tiny} {map-} \end{tiny}};
\node[rotate=18,font=\sffamily,mycolor5] at (9,20.7) {\begin{tiny} {based} \end{tiny}};
\node[rotate=10,font=\sffamily,mycolor6] at (8.3,8.7) {\begin{tiny} {interactive} \end{tiny}};
\addplot [color=mycolor4]
  table[row sep=crcr]{%
0	0.642464318101722\\
0.5	1.11617657444028\\
1	1.83333901109299\\
1.5	2.82216086540523\\
2	4.04622469953553\\
2.5	5.52187077024517\\
3	7.2332093927498\\
3.5	9.19431297543774\\
4	11.2685090546882\\
4.5	13.4725500354325\\
5	15.7464314484639\\
5.5	18.2283986509136\\
6	20.854123615825\\
6.5	23.5432279934706\\
7	26.4648739357112\\
7.5	29.4422704593718\\
8	32.5918690398899\\
8.5	35.9692206479254\\
9	39.4273967707703\\
9.5	43.031365006759\\
10	47.2504742102223\\
};

\addplot [color=mycolor5]
  table[row sep=crcr]{%
0	1.90415792938544\\
0.5	2.89008165872203\\
1	4.05475405949689\\
1.5	5.35086271663593\\
2	6.69039423073931\\
2.5	8.11780704116135\\
3	9.61521575148154\\
3.5	11.0033320987247\\
4	12.1559529903745\\
4.5	13.1895290881139\\
5	13.9937296469477\\
5.5	15.0621809439645\\
6	16.2082610344765\\
6.5	17.3283051724111\\
7	18.8322769713507\\
7.5	20.2077847834465\\
8	21.802694978443\\
8.5	23.3257366940292\\
9	24.9910329306401\\
9.5	26.6477676471905\\
10	28.0549512897815\\
};

\addplot [color=mycolor6]
  table[row sep=crcr]{%
0	0.655777505881241\\
0.5	1.10345510029177\\
1	1.74932765198259\\
1.5	2.58206079435291\\
2	3.56624627890409\\
2.5	4.69974815331098\\
3	5.98667149740603\\
3.5	7.06047310315904\\
4	7.75455924851636\\
4.5	8.26118752979944\\
5	8.2676821105101\\
5.5	8.76665340845543\\
6	9.35263689112036\\
6.5	9.69574605791548\\
7	10.8995131934371\\
7.5	11.5283838380895\\
8	12.5869015551959\\
8.5	13.4063323461425\\
9	14.61058151107\\
9.5	15.9164186568161\\
10	16.9525108677093\\
};

\end{axis}
\end{tikzpicture}%
\caption{Detailed evaluation of agent $\agent^0$ in scene~1 (first row, simulated data) and scene~2 (second row, real data): Comparison of route estimation and trajectory prediction for the different tracking methods (CTRV, map-based, interactive), also showing the scaled velocity profile $\tilde{v}=v/(50\mathrm{km/h})$.
}
\label{fig:eval_scenes}
\vspace{-9px}
\end{figure*}

In order to assess the necessity of interaction-aware prediction, we compare our model to simpler models in simulated and real driving scenarios.
Scenes with interactive behavior are recorded with a measuring vehicle on real roads and on a test-track and are generated with a proprietary traffic simulator. The measuring vehicle's pose and velocity is estimated using GPS/INS. Both lidar and radar sensors are used to detect and track objects nearby.
The evaluation parameters can be seen in \Fref{tab:eval_param}. 
To avoid particle deprivation due to resampling, new particles are sampled from the current measurement distribution with probability $0.001$.
The computing time of one time step of a scene with three vehicles, each having three route options, is approximately $\SI{0.3}{\second}$ on an Intel Core i7-5820K CPU @ 3.30GHz with non-optimized C++ code.
\mytodo{For the control of the agents, a proprietary simulator is used.}
\begin{table}
      \vspace{5px}
\begin{center}
\setlength{\tabcolsep}{.63em}
\caption{Evaluation Parameters}
\label{tab:eval_param}
\begin{tabular}{llcllcll}
\toprule
$\Delta T$&\SI{0.2}{\second}&&
$\delta$&4&&
$\sigma_{z_{x/y}}$&\SI{15}{\meter}\\

$N$&1000&&
$a_\mathrm{lat}^\mathrm{max}$&\SI{2}{\meter\per\square\second}&&
$\sigma_{z_\theta}$&\SI{3.14}{}\\

$l_\mathrm{H}$&\SI{30}{\meter}&&
$\sigma_{x/y}$&\SI{0.5}{\meter}&&
$\sigma_{z_v}$&\SI{15}{\meter\per\second}\\

$d_d$&\SI{2}{\meter}&&
$\sigma_{\theta}$&\SI{0.05}{}&&
$\sigma_{s_{x/y}}$&\SI{1}{\meter}\\

$T_d$&\SI{0.1}{\second}&&
$\sigma_{v}$&\SI{1.5}{\meter\per\second}&&
$\sigma_{s_\theta}$&\SI{0.03}{}\\

$a_d$&\SI{0.7}{\meter\per\square\second}&&
$\sigma_{a}$&\SI{1.5}{\meter\per\square\second}&&
$\sigma_{s_v}$&\SI{1}{\meter\per\second}\\

$b_d$&\SI{-0.5}{\meter\per\square\second}&&
$\sigma_{\dot{\theta}}$&\SI{0.05}{\per\second}&&
\\

\bottomrule
\end{tabular}
\end{center}
      \vspace{-9px}
\end{table}

\mytodo{add all relevant values}

\subsection{Intention Estimation}

The imprecision of the route (and analogously maneuver) estimate is measured using the Kullback-Leibler divergence  
\begin{align}
D_{\mathrm{KL}}(r^i_\textrm{GT}\|r^i) = \sum_{j=1}^{|R|} r^i_{\textrm{GT},j} \, \log\frac{r^i_{\textrm{GT},j}}{P(r^i_j)}
\end{align}
from estimate $r^i=[P(r^i_1), \cdots, P(r^i_{|R|})]$ to ground truth $r^i_\textrm{GT}= [r^i_{\mathrm{GT},1}, \cdots, r^i_{\mathrm{GT},|R|}]$, with
\begin{align}
r^i_{\mathrm{GT},j} = 
\begin{cases} 
1 & \textrm{if }\agent^i \textrm{ follows } r^i_j  \\
0 & \textrm{else} \\
\end{cases}.
\end{align}

We evaluate the intention estimation of the presented model, which we call \emph{interactive} model, and a solely \emph{map-based} model.
The \emph{map-based} model uses all of the features given by the map but ignores surrounding vehicles and, therefore, is interaction-unaware. Thus, agents are predicted as if there were no other vehicles around.

As our dataset mostly consists of scenes with little interaction and both models are identical for scenes without interaction, a statistical evaluation of the complete dataset produces similar results.
In order to highlight their differences, we specifically determined situations in which multiple vehicles cross an intersection, and hence, containing interdependencies between vehicles. Though intersection crossings are statistically rare in our dataset, these situations tend to be most critical and therefore require explicit evaluation.
The intention estimation is evaluated in detail for three scenes:

\emph{1) Yielding vehicle:}
In the simulated \emph{scene~1} (first row of \Fref{fig:eval_scenes}), $\agent^1$ has right of way and goes straight, $\agent^0$ has to yield and wants to turn left. 
To improve readability, at first it is assumed that $\agent^0$ is actually yielding and therefore only has one possible maneuver ($\agent^1 {\prec} \agent^0$), but multiple possible routes. 
While $\agent^0$'s routes for going straight and turning left demand yielding, the route for turning right is free. As $\agent^0$ waits for $\agent^1$ ($t{=}\SI{10}{}{-}\SI{18}{\second}$), it is inferred by the interactive model that turning right is unlikely (as waiting would not be necessary) and turning left and going straight is equally likely (as both routes are blocked). As soon as $\agent^1$ has left the conflict area, $\agent^0$ accelerates again and turns, whereby the left route is inferred correctly.
The map-based model, however, infers incorrectly that $\agent^0$ wants to turn right ($t{=}\SI{13}{\second}$), as this route has the highest curvature, implying the lowest velocity. For $t{>}\SI{13}{\second}$, as $\agent^0$ even becomes too slow for turning right, none of the map-based models can explain the actual behavior anymore. Thus, only the particles sampled newly from the measurement survive, resulting in a random oscillation and a momentary improvement of the $D_\mathrm{KL}$.

\emph{1b) Maneuver distinction:}
The combined maneuver and route estimation is analyzed in \Fref{fig:eval_maneuver_prob}, where scene~1 is modified, such that $\agent^0$ crosses first (scene~1b).
The interactive model with maneuver distinction is compared to the interactive model without maneuver distinction (assuming $\agent^0$ will yield):
At ${t{=}\SI{0}{\second}}$, all routes are equally likely, but as $\agent^0$  does not decelerate strongly (${t{=}\SI{2}{}{-}\SI{10}{\second}}$), the probability to yield decreases, whereas the probabilities to either turn right (no conflict) or merge / cross before $\agent^1$ increase. 
As $\agent^0$ slows down in order to respect the upcoming curvature (${t{=}\SI{9}{}{-}\SI{11}{\second}}$), the straight route becomes unlikely. 
Finally (${t{=}\SI{12}{}{-}\SI{20}{\second}}$), as the velocity is still too high for turning right, it is correctly inferred that $\agent^0$ will turn left and merge before $\agent^1$.
Without the distinction of the two possible maneuvers, assuming $\agent^0$ is going to yield, it is incorrectly inferred that $\agent^0$ wants to turn right (as this lane has no conflict), resulting in a higher estimation and trajectory prediction error.

\emph{2) Preceding vehicle:}
In the real driving \emph{scene~2} (second row of \Fref{fig:eval_scenes}), $\agent^0$ follows $\agent^1$ approaching an intersection. As $\agent^1$ has to yield and therefore decelerates, $\agent^0$ decelerates as well in order to keep the desired headway distance.
All three possible routes of $\agent^0$ are blocked by the preceding vehicle, hence, it is not possible to infer the route until the preceding agent has passed the intersection ($t{=}\SI{17}{\second}$). A uniform distribution is the desired result, which is generated by the interactive method. The map-based method incorrectly infers that $\agent^0$ wants to turn right (${t{=}\SI{10}{\second}}$), as it is slowing down (actually caused by the preceding vehicle). For ${t{>}\SI{10}{\second}}$, none of the map-based models can explain the observations anymore, also resulting in a random oscillation.

\renewcommand{\mywidth}{1.2in}
\renewcommand{\myheight}{0.5in}

\begin{figure}
\centering
\footnotesize
  \input{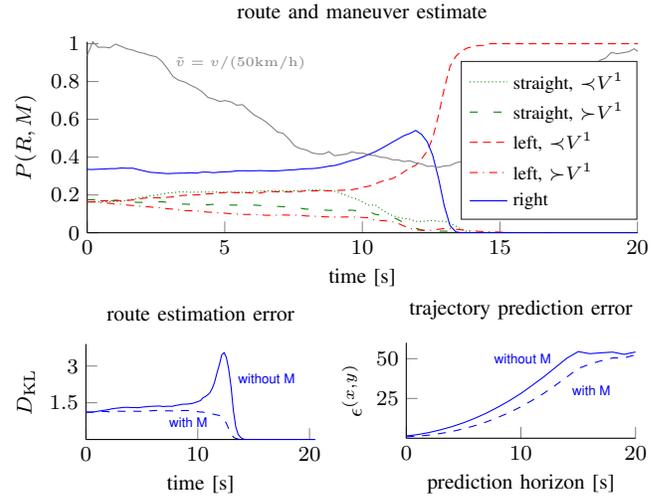}
%
\begin{tikzpicture}

\begin{axis}[%
width=\mywidth,
height=\myheight,
at={(0.753in,0.484in)},
scale only axis,
xmin=0,
xmax=20.51,
ymin=0,
xlabel={time [s]},
ylabel={$D_\mathrm{KL}$},
axis background/.style={fill=white},
title={route estimation error},
axis x line*=bottom,
axis y line*=left,
ytick={1.5, 3},
xtick={0, 10, 20},
legend style={legend cell align=left, align=left, draw=white!15!black}
]
\node[font=\sffamily,mycolor6] at (16,2.5) {\begin{tiny} {without M} \end{tiny}};
\node[font=\sffamily,mycolor6] at (9,0.7) {\begin{tiny} {with M} \end{tiny}};
\addplot [color=mycolor6, dashed]
  table[row sep=crcr]{%
0	1.11382745447456\\
0.2	1.11718367425361\\
0.4	1.1150465951316\\
0.61	1.11565672327743\\
0.82	1.12393009665246\\
1.03	1.12639466968062\\
1.24	1.13724917228309\\
1.45	1.14413280396998\\
1.66	1.14507516307851\\
1.87	1.14005947858235\\
2.08	1.13010295575954\\
2.29	1.13444674167739\\
2.5	1.1403722229164\\
2.71	1.1507480430887\\
2.92	1.15391360445326\\
3.13	1.15264617716066\\
3.34	1.14948461904193\\
3.55	1.15296288341825\\
3.76	1.14790758039305\\
3.97	1.14759247088989\\
4.18	1.15772555316068\\
4.39	1.15327969001048\\
4.5999999999999	1.15264617716066\\
4.8099999999999	1.15232957117416\\
5.0199999999999	1.1510641486706\\
5.2299999999999	1.15677120284171\\
5.4399999999999	1.1666770195823\\
5.6499999999999	1.1708604528758\\
5.8599999999999	1.17279718685755\\
6.0699999999999	1.1815592094891\\
6.2799999999999	1.18417017702982\\
6.4899999999999	1.17441400208446\\
6.6999999999999	1.16731948667796\\
6.9099999999999	1.17182835099943\\
7.1199999999999	1.1689274625832\\
7.3299999999999	1.1715056141885\\
7.5399999999999	1.16989349067314\\
7.7499999999999	1.16699820153464\\
7.9599999999999	1.16635594075469\\
8.1699999999999	1.17312034084324\\
8.3799999999999	1.17376696226893\\
8.5899999999999	1.17928019173563\\
8.7899999999999	1.18482398548724\\
8.9899999999999	1.18384343303084\\
9.1999999999998	1.17928019173563\\
9.4099999999998	1.1715056141885\\
9.6199999999998	1.16764087507861\\
9.8299999999998	1.15138035420683\\
10.04	1.13227249016081\\
10.25	1.11963165589225\\
10.46	1.12177856670001\\
10.67	1.09034654425114\\
10.88	1.07265218699864\\
11.09	1.06016110229391\\
11.3	1.04639941722123\\
11.51	1.02499014879755\\
11.73	1.00021596681197\\
11.94	0.98323213762809\\
12.15	0.907578791272208\\
12.36	0.809680996815964\\
12.56	0.659712404473785\\
12.77	0.442232832300457\\
12.98	0.238891908282438\\
13.19	0.0994888985008309\\
13.4	0.046043938501501\\
13.61	0.0202027073176113\\
13.82	0.00682322534821607\\
14.03	0.00200200267076338\\
14.23	0.000500125041772256\\
14.43	9.01501095995586e-14\\
14.64	0.000100005000423417\\
14.85	0.000100005000422306\\
15.06	0.000100005000413204\\
15.27	0.000100005000413204\\
15.48	8.99280649946336e-14\\
15.69	9.01501095995586e-14\\
15.9	9.01501095995586e-14\\
16.11	9.01501095995586e-14\\
16.32	9.01501095995586e-14\\
16.53	0.000100005000423417\\
16.74	0.000200020002757155\\
16.95	9.01501095995586e-14\\
17.16	9.01501095995586e-14\\
17.37	9.01501095995586e-14\\
17.58	9.01501095995586e-14\\
17.8	9.01501095995586e-14\\
18.01	9.01501095995586e-14\\
18.22	9.01501095995586e-14\\
18.42	9.01501095995586e-14\\
18.63	0.000100005000423417\\
18.84	9.01501095995586e-14\\
19.05	9.01501095995586e-14\\
19.26	9.01501095995586e-14\\
19.47	9.01501095995586e-14\\
19.68	9.01501095995586e-14\\
19.88	9.01501095995586e-14\\
20.09	0.000100005000423417\\
20.3	9.01501095995586e-14\\
20.51	9.01501095995586e-14\\
};

\addplot [color=mycolor6]
  table[row sep=crcr]{%
0	1.11382745447456\\
0.2	1.11963165589225\\
0.4	1.11748934443786\\
0.61	1.12423783630714\\
0.82	1.13134211329109\\
1.03	1.15043203739797\\
1.24	1.16475209117272\\
1.45	1.18123331735651\\
1.66	1.19303286428767\\
1.87	1.20999087665157\\
2.08	1.22417551164352\\
2.29	1.25071506687814\\
2.5	1.26584820804406\\
2.71	1.2761851384399\\
2.92	1.28409884880526\\
3.13	1.29755115128554\\
3.34	1.31118688863312\\
3.55	1.31118688863312\\
3.76	1.31453199489304\\
3.97	1.30305676254987\\
4.18	1.30453005931729\\
4.39	1.30563645810247\\
4.5999999999999	1.29681935469232\\
4.8099999999999	1.30526752247755\\
5.0199999999999	1.3208812225717\\
5.2299999999999	1.33142745967909\\
5.4399999999999	1.34055646944607\\
5.6499999999999	1.35208618979019\\
5.8599999999999	1.35402080056963\\
6.0699999999999	1.35712398382015\\
6.2799999999999	1.3621872857767\\
6.4899999999999	1.36609965383441\\
6.6999999999999	1.36179688951954\\
6.9099999999999	1.34400144892968\\
7.1199999999999	1.35363357888038\\
7.3299999999999	1.35673555887838\\
7.5399999999999	1.35790128661871\\
7.7499999999999	1.35324650707373\\
7.9599999999999	1.38629436111993\\
8.1699999999999	1.39958225544687\\
8.3799999999999	1.41716602478702\\
8.5899999999999	1.43674590461603\\
8.7899999999999	1.44477129087891\\
8.9899999999999	1.44180707105019\\
9.1999999999998	1.45457320187323\\
9.4099999999998	1.46577056262173\\
9.6199999999998	1.48501033405909\\
9.8299999999998	1.52096926444654\\
10.04	1.55211284581488\\
10.25	1.59061615819356\\
10.46	1.70045731082127\\
10.67	1.75100147675589\\
10.88	1.81892513730082\\
11.09	1.91324936681576\\
11.3	2.08908791873164\\
11.51	2.34132592131046\\
11.73	2.7318307297676\\
11.94	3.0597376035299\\
12.15	3.46094738606793\\
12.36	3.54391368386375\\
12.56	3.35526844977376\\
12.77	2.87528612047811\\
12.98	2.1663074747015\\
13.19	1.17993080984576\\
13.4	0.523404405972935\\
13.61	0.216292076365195\\
13.82	0.0915671935255782\\
14.03	0.0283994745217905\\
14.23	0.00471107973020967\\
14.43	0.000500125041772256\\
14.64	9.01501095995586e-14\\
14.85	9.01501095995586e-14\\
15.06	8.01581023779331e-14\\
15.27	7.99360577730081e-14\\
15.48	8.90398865749336e-14\\
15.69	9.01501095995586e-14\\
15.9	9.01501095995586e-14\\
16.11	9.01501095995586e-14\\
16.32	9.01501095995586e-14\\
16.53	9.01501095995586e-14\\
16.74	9.01501095995586e-14\\
16.95	9.01501095995586e-14\\
17.16	9.01501095995586e-14\\
17.37	0.000100005000423417\\
17.58	9.01501095995586e-14\\
17.8	9.01501095995586e-14\\
18.01	9.01501095995586e-14\\
18.22	9.01501095995586e-14\\
18.42	9.01501095995586e-14\\
18.63	9.01501095995586e-14\\
18.84	9.01501095995586e-14\\
19.05	9.01501095995586e-14\\
19.26	9.01501095995586e-14\\
19.47	9.01501095995586e-14\\
19.68	9.01501095995586e-14\\
19.88	9.01501095995586e-14\\
20.09	9.01501095995586e-14\\
20.3	9.01501095995586e-14\\
20.51	9.01501095995586e-14\\
};

\end{axis}
\end{tikzpicture}%
%
%
%
\begin{tikzpicture}

\begin{axis}[%
width=\mywidth,
height=\myheight,
at={(0.78in,0.521in)},
scale only axis,
xmin=0,
xmax=20,
ymin=0,
ymax=60,
xlabel={prediction horizon [s]},
ylabel={$\epsilon^{(x,y)}$},
axis background/.style={fill=white},
title={trajectory prediction error},
axis x line*=bottom,
axis y line*=left,
ytick={25, 50},
xtick={0, 10, 20},
legend style={at={(0.349,0.819)}, anchor=south west, legend cell align=left, align=left, draw=white!15!black}
]
\node[font=\sffamily,mycolor6] at (16,30) {\begin{tiny} {with M} \end{tiny}};
\node[font=\sffamily,mycolor6] at (10,50) {\begin{tiny} {without M} \end{tiny}};
\addplot [color=mycolor6, dashed]
  table[row sep=crcr]{%
0	0.951631823263218\\
1	1.31522090746098\\
2	1.99177719577585\\
3	2.99015613227312\\
4	4.29101340885682\\
5	5.91761338005136\\
6	7.93074247620359\\
7	10.2769449759541\\
8	13.080985446343\\
9	16.2106981204552\\
10	19.7440426540796\\
11	23.8548992891662\\
12	28.2406699428208\\
13	33.2506717591852\\
14	38.4551178450501\\
15	43.8012708561907\\
16	46.5745109331638\\
17	48.6893973874829\\
18	50.4671345804694\\
19	50.483890974724\\
20	52.4252956131113\\
};

\addplot [color=mycolor6]
  table[row sep=crcr]{%
0	1.40581480998787\\
1	2.2960182216145\\
2	3.61605826567155\\
3	5.30567112651752\\
4	7.33024878707053\\
5	9.72992360976508\\
6	12.6020376037884\\
7	15.8170424702126\\
8	19.5600359503191\\
9	23.6356771598594\\
10	28.1549004287085\\
11	33.3282391904586\\
12	38.6298789143698\\
13	44.4334542614771\\
14	50.2850662862347\\
15	54.4356186219021\\
16	53.1382357421854\\
17	53.6731000365056\\
18	54.1252475520425\\
19	52.7135141885868\\
20	54.188189185821\\
};

\end{axis}
\end{tikzpicture}%
\caption{Route and maneuver estimation of agent $\agent^0$ in scene~1b and comparison of interaction-aware model with and without maneuver distinction.}  
\label{fig:eval_maneuver_prob}  
\vspace{-9px}  
\end{figure}

\mytodo{
Scenes for statistical evaluation:
-1s2l
-1r2s_follow
-
}

\mytodo{
ROUTE ESTIMATION: map-based vs. fully vs. partly
specific example (one scene):
-s=0,n=high: route estimation performance for three models: KLD or similar
-s=var, n=var: complexity comparison for three models (only two without map-based??): variance plots (similar to boxplots) depending on number of particles
statistics (multiple scenes)
-s=0,n=high: route estimation performance for three models: KLD or similar
}

\mytodo{Konvergenzkriterium überlegen, prüfen wie viele Partikel benötigt}

\subsection{Trajectory Prediction}

The accuracy of the trajectory prediction of all agents at time $t$ for the future time step $\tau$ is quantified using the position components of the weighted root mean square error between prediction and measurement
\begin{align}
\mathrm{\epsilon^{(x,y)}_{\tau|t}} = \sqrt{\sum_{R_{t},M_{t}}{P(R_{t},M_{t}) \left(\hat{X}^{(x,y)}_{\tau|t,R_t,M_t} - Z^{(x,y)}_{\tau}\right)^2}},
\end{align}
and the measurement likelihood
\begin{align}
\mathrm{\mathcal{L}^{(x,y)}_{\tau|t}} = {\prod_{R_t,M_t}{P(R_t,M_t)~P( Z^{(x,y)}_{\tau} | \hat{X}^{(x,y)}_{\tau|t,R_t,M_t})}}.
\end{align}
\mytodo{use RWSE from Wheeler et al instead of RMSE??}

The \emph{interactive} model is compared to the \emph{map-based} model and a \emph{\gls{ctrv}} model~\cite{schubert_comparison_2008}, which serves as a simple baseline algorithm.
It is independent of both the map and surrounding vehicles.
The error of the trajectory prediction of $\agent^0$ for scenes~1 and 2 are depicted in the most right column of \Fref{fig:eval_scenes}. The CTRV model performs worse in scene~1, as $\agent^0$ changes its velocity and orientation more intensely. For the map-based model, the first scene is also more challenging, as $\agent^0$ stops for a long time, which cannot be explained by the model at all. Its high route estimation error negatively affects its prediction accuracy. The interactive model outperforms the other two approaches in both scenes.

\begin{figure}
      \vspace{4px}
	\centering
	\footnotesize
	\includegraphics[scale=0.136, trim=0 350 400 70, clip=true]{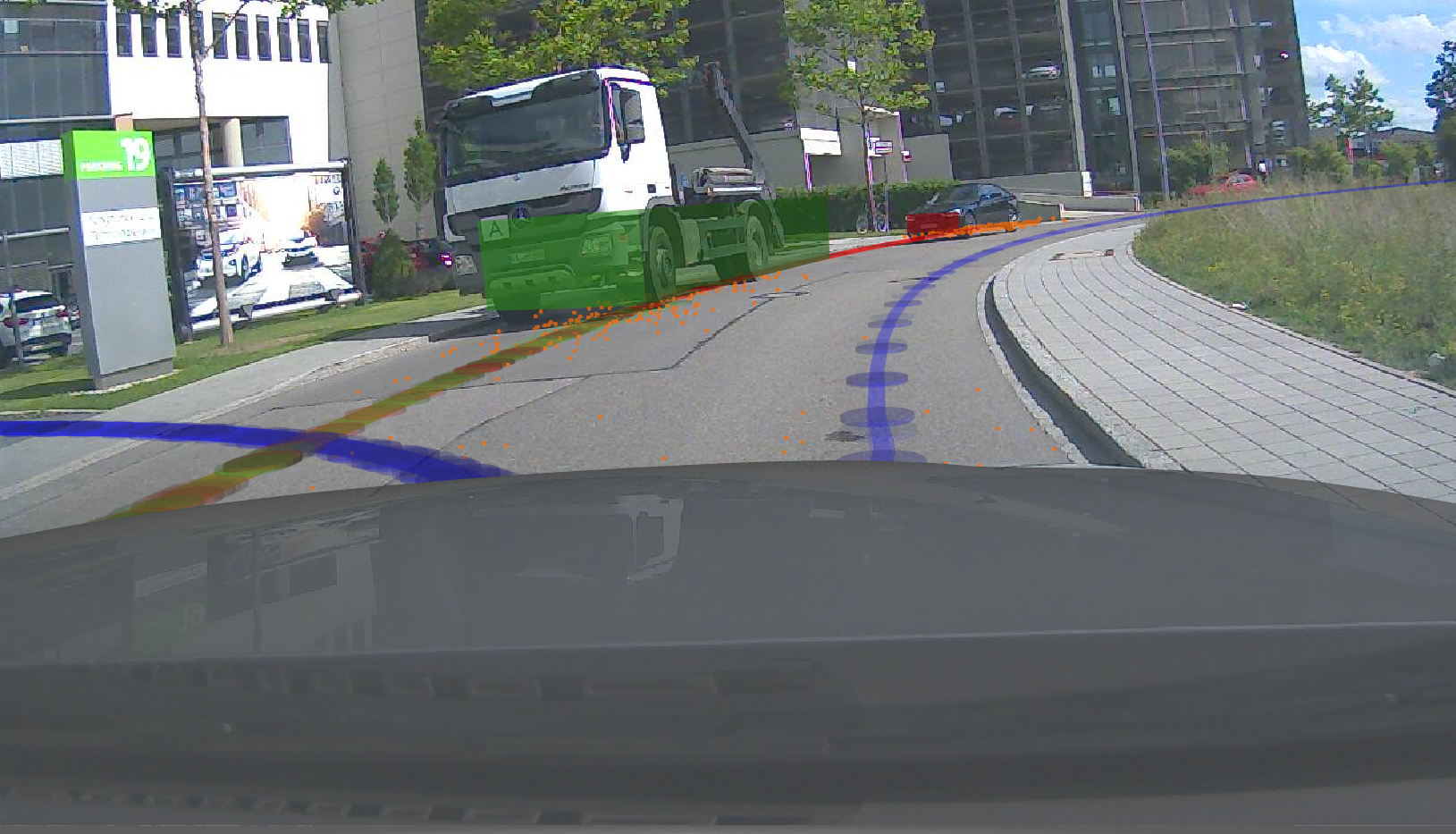}
	\caption{Camera view of measurement vehicle while yielding to oncoming traffic in order to turn left into a parking lot.}
	\label{fig:eval_parkring}
	\vspace{-15px}
\end{figure}

Furthermore, in order to compare the models in a quantitative manner, five different real driving scenes have been recorded on a test track and on real roads (\Fref{fig:eval_parkring}). These scenes altogether consist of 15 vehicles, two four-way intersections, two T-junctions, and a roundabout.
The statistical results showing the prediction error and measurement likelihood over all scenes and vehicles are depicted in \Fref{fig:eval_real_statistical}.
It can be seen that the interaction-aware model outperforms both CTRV and map-based models.
Although the differences between the map-based and the interactive model might seem to be rather small, it has to be noted that the time steps in which traffic participants actually interact with each other do not predominate. 
As shown in \Fref{fig:eval_scenes}, however, in scenes where the behaviors of drivers are highly interdependent, interaction-aware prediction becomes essential.

A video of the approach with exemplary scenes is included in the conference proceedings and can also be found at \mbox{\url{https://mediatum.ub.tum.de/1449806}}.

\renewcommand{\mywidth}{1.2in}
\renewcommand{\myheight}{0.7in}
	
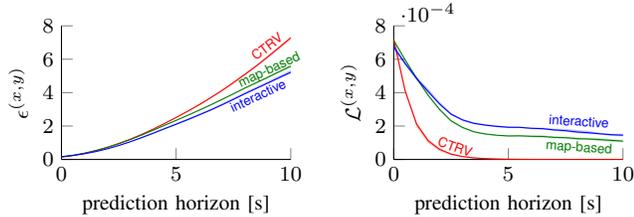
\begin{figure}
	\vspace{5px}
	\centering
	\footnotesize
	\begin{subfigure}{0.49\textwidth} 
		\centering
	\footnotesize
%
%
%
\begin{tikzpicture}

\begin{axis}[%
width=\mywidth,
height=\myheight,
at={(0.766in,0.486in)},
scale only axis,
xmin=0,
xmax=10,
ymin=0,
ymax=8,
xlabel={prediction horizon [s]},
ylabel={$\epsilon^{(x,y)}$},
axis background/.style={fill=white},
title style={font=\bfseries},
axis x line*=bottom,
axis y line*=left,
clip=false,
xtick={0, 5, 10}
]
\node[rotate=37,font=\sffamily,mycolor4] at (8.9,6.7) {\begin{tiny} {CTRV} \end{tiny}};
\node[rotate=25,font=\sffamily,mycolor5] at (9.,5.3) {\begin{tiny} {map-based} \end{tiny}};
\node[rotate=25,font=\sffamily,mycolor6] at (8.5,3.8) {\begin{tiny} {interactive} \end{tiny}};
\addplot [color=mycolor4, forget plot]
  table[row sep=crcr]{%
0	0.138249995350613\\
0.5	0.239228959218719\\
1	0.370713319661487\\
1.5	0.534737657051238\\
2	0.730606619106832\\
2.5	0.957769675270857\\
3	1.21324298196795\\
3.5	1.49940012335742\\
4	1.81403371355763\\
4.5	2.15720363238821\\
5	2.50947232732912\\
5.5	2.88258681110404\\
6	3.26934890990864\\
6.5	3.68956891775578\\
7	4.12617190668678\\
7.5	4.59953580403891\\
8	5.08900257873654\\
8.5	5.59877228531907\\
9	6.14514798763328\\
9.5	6.7078601730246\\
10	7.27812947552921\\
};

\addplot[area legend, draw=none, fill=mycolor4, fill opacity=0.25, forget plot]
table[row sep=crcr] {%
x	y\\
0	0.143888244179241\\
0.5	0.24647977320912\\
1	0.380032247764142\\
1.5	0.546222532502232\\
2	0.745184795893078\\
2.5	0.975410979967769\\
3	1.23385914069314\\
3.5	1.52191673847414\\
4	1.8383304362524\\
4.5	2.18649738188922\\
5	2.54681712593766\\
5.5	2.92992028194971\\
6	3.32590001744835\\
6.5	3.75195133920508\\
7	4.19470710761797\\
7.5	4.67255594771275\\
8	5.16583226329772\\
8.5	5.67813582302214\\
9	6.22692037561172\\
9.5	6.78842410091661\\
10	7.34900172990744\\
10	7.17383076073053\\
9.5	6.61546333371465\\
9	6.06403743007585\\
8.5	5.5311239543324\\
8	5.03255073958545\\
7.5	4.54194729580119\\
7	4.07300942401887\\
6.5	3.63959145283979\\
6	3.22384331394865\\
5.5	2.841470695425\\
5	2.47303301519006\\
4.5	2.12717093436974\\
4	1.79086424964986\\
3.5	1.48402977124262\\
3	1.2012601191896\\
2.5	0.945286841612499\\
2	0.718474857721663\\
1.5	0.525597626775466\\
1	0.364292948282896\\
0.5	0.234603592786499\\
0	0.134436568031181\\
}--cycle;
\addplot [color=mycolor5, forget plot]
  table[row sep=crcr]{%
0	0.16495446126997\\
0.5	0.248945111780235\\
1	0.371040609411695\\
1.5	0.527544130495097\\
2	0.715826078511458\\
2.5	0.939617621777082\\
3	1.18760056054764\\
3.5	1.46271769531277\\
4	1.74954445503008\\
4.5	2.04606680780129\\
5	2.33925988128093\\
5.5	2.63809557993923\\
6	2.94639615379385\\
6.5	3.25959666500633\\
7	3.59238892215829\\
7.5	3.92900154043187\\
8	4.27492009623386\\
8.5	4.5951952406019\\
9	4.92137539003382\\
9.5	5.25746203516784\\
10	5.58758332791197\\
};

\addplot[area legend, draw=none, fill=mycolor5, fill opacity=0.25, forget plot]
table[row sep=crcr] {%
x	y\\
0	0.166571480782627\\
0.5	0.250655202356005\\
1	0.373262307203978\\
1.5	0.530918718732967\\
2	0.720764412074052\\
2.5	0.947132194306063\\
3	1.1982341434462\\
3.5	1.4766004758786\\
4	1.76728824763329\\
4.5	2.06899213847112\\
5	2.36548279599129\\
5.5	2.66844269161587\\
6	2.98086724674633\\
6.5	3.29397296627391\\
7	3.62755532323637\\
7.5	3.96143313628871\\
8	4.30578335922808\\
8.5	4.63043772095537\\
9	4.96037947213206\\
9.5	5.29777829124241\\
10	5.63201298355479\\
10	5.50466938933127\\
9.5	5.18310744614603\\
9	4.85713745890516\\
8.5	4.5395269648152\\
8	4.23097441783705\\
7.5	3.89697972090369\\
7	3.56484767077623\\
6.5	3.22919745733995\\
6	2.91277260195484\\
5.5	2.60293368917662\\
5	2.30537584798755\\
4.5	2.01359971838329\\
4	1.72177567664175\\
3.5	1.44147953989158\\
3	1.17219456547171\\
2.5	0.928114969073579\\
2	0.707697116448812\\
1.5	0.52224633425293\\
1	0.367785345798809\\
0.5	0.247080372441134\\
0	0.163135834295777\\
}--cycle;
\addplot [color=mycolor6, forget plot]
  table[row sep=crcr]{%
0	0.164193466845417\\
0.5	0.235842601975472\\
1	0.33927459883738\\
1.5	0.471801807958932\\
2	0.63578584569349\\
2.5	0.834851928929114\\
3	1.05767660157569\\
3.5	1.30717383309284\\
4	1.56978061699779\\
4.5	1.84197556452647\\
5	2.10775873804197\\
5.5	2.37931384326548\\
6	2.66405463307318\\
6.5	2.95528616878332\\
7	3.26895389257673\\
7.5	3.59582908847404\\
8	3.93508781446509\\
8.5	4.25363846821951\\
9	4.575145928729\\
9.5	4.90062002948223\\
10	5.21840149162588\\
};

\addplot[area legend, draw=none, fill=mycolor6, fill opacity=0.25, forget plot]
table[row sep=crcr] {%
x	y\\
0	0.169439235802298\\
0.5	0.244807567023164\\
1	0.353518791653548\\
1.5	0.491678417052244\\
2	0.661664732751586\\
2.5	0.867790191876575\\
3	1.09790704410891\\
3.5	1.35925185707639\\
4	1.62547180226676\\
4.5	1.89909326022848\\
5	2.17563463043232\\
5.5	2.44831274634201\\
6	2.73453291678929\\
6.5	3.03225132424065\\
7	3.3545219193044\\
7.5	3.68976094552373\\
8	4.0292756602232\\
8.5	4.34817974713266\\
9	4.67743349597069\\
9.5	5.019169801713\\
10	5.3445436008539\\
10	5.13725522594499\\
9.5	4.83678720572334\\
9	4.5236023971947\\
8.5	4.21054310114756\\
8	3.8944719401883\\
7.5	3.55266753965498\\
7	3.23005056682607\\
6.5	2.90719921852201\\
6	2.61443850871897\\
5.5	2.32417461531676\\
5	2.05234149171178\\
4.5	1.7919528057377\\
4	1.52042023551797\\
3.5	1.26700699316328\\
3	1.02627420272273\\
2.5	0.810801282704457\\
2	0.618494640890611\\
1.5	0.46036994604606\\
1	0.332421210928155\\
0.5	0.231659829457863\\
0	0.160812799795613\\
}--cycle;
\end{axis}
\end{tikzpicture}%
\hspace{0.2cm}
		\centering
	\footnotesize
%
%
%
\begin{tikzpicture}

\begin{axis}[%
width=\mywidth,
height=\myheight,
at={(0.78in,0.521in)},
scale only axis,
xmin=0,
xmax=10,
ymin=0,
ymax=0.0008,
xlabel={prediction horizon [s]},
ylabel={$\mathcal{L}^{{(x,y)}}$},
axis background/.style={fill=white},
title style={font=\bfseries},
axis x line*=bottom,
axis y line*=left,
xtick={0, 5, 10},
legend style={legend cell align=left, align=left, draw=white!15!black, font=\scriptsize}
]
\node[rotate=-20,font=\sffamily,mycolor4] at (2.6,0.00008) {\begin{tiny} {CTRV} \end{tiny}};
\node[rotate=-4,font=\sffamily,mycolor5] at (8,0.00007) {\begin{tiny} {map-based} \end{tiny}};
\node[rotate=-4,font=\sffamily,mycolor6] at (8,0.00023) {\begin{tiny} {interactive} \end{tiny}};
\addplot [color=mycolor4]
  table[row sep=crcr]{%
0	0.000715730728476356\\
0.5	0.000410038141161607\\
1	0.000211264964130854\\
1.5	0.000112131019455682\\
2	6.02289587769508e-05\\
2.5	3.22306158941793e-05\\
3	1.72269250778363e-05\\
3.5	9.48283229205517e-06\\
4	5.1219601788578e-06\\
4.5	2.85941912738314e-06\\
5	1.65811600169506e-06\\
5.5	1.02872378230885e-06\\
6	6.59229193535724e-07\\
6.5	3.94943416601244e-07\\
7	2.24698066125947e-07\\
7.5	1.2739880503799e-07\\
8	8.59606473207431e-08\\
8.5	6.24737882245158e-08\\
9	4.01654382769312e-08\\
9.5	2.11286116862588e-08\\
10	5.02326618888473e-09\\
};

\addplot[area legend, draw=none, fill=mycolor4, fill opacity=0.25, forget plot]
table[row sep=crcr] {%
x	y\\
0	0.000746151625429833\\
0.5	0.000424463299431778\\
1	0.000223576634209046\\
1.5	0.000123269617269679\\
2	6.83380694936908e-05\\
2.5	3.70186487537169e-05\\
3	2.1673407618062e-05\\
3.5	1.30632374035082e-05\\
4	7.72867604322088e-06\\
4.5	4.77193352893096e-06\\
5	2.99321665000654e-06\\
5.5	1.84383867430862e-06\\
6	1.64270670298423e-06\\
6.5	9.79373148329337e-07\\
7	5.54143554048525e-07\\
7.5	3.896051125311e-07\\
8	3.18198610744188e-07\\
8.5	2.45805475618156e-07\\
9	1.0780994845596e-07\\
9.5	4.74121694646854e-08\\
10	1.31433945782932e-08\\
10	7.7553764452935e-10\\
9.5	1.13804916973898e-09\\
9	3.29074617608636e-09\\
8.5	4.06922892048556e-09\\
8	1.06760868526582e-08\\
7.5	2.37128675892656e-08\\
7	5.08679472461253e-08\\
6.5	1.18941610141681e-07\\
6	2.35550732661785e-07\\
5.5	4.81709501271116e-07\\
5	8.73718501673987e-07\\
4.5	1.53974740445459e-06\\
4	3.1853376455217e-06\\
3.5	6.64109307858369e-06\\
3	1.28591918923584e-05\\
2.5	2.55236726168089e-05\\
2	5.06886486162865e-05\\
1.5	9.81511647031861e-05\\
1	0.000192634984749431\\
0.5	0.000398069553024778\\
0	0.000700849942799344\\
}--cycle;
\addplot [color=mycolor5]
  table[row sep=crcr]{%
0	0.000711539444634911\\
0.5	0.000595700031568474\\
1	0.000486599018393359\\
1.5	0.000385027213549657\\
2	0.000299786289479956\\
2.5	0.000231715667706977\\
3	0.000191773572687586\\
3.5	0.000165840296373593\\
4	0.000152520562174954\\
4.5	0.000145004022204195\\
5	0.000140809372539622\\
5.5	0.000141523511234907\\
6	0.000138581379233568\\
6.5	0.000137480540904033\\
7	0.000132216880739004\\
7.5	0.000130273600645164\\
8	0.000124623467944928\\
8.5	0.000122318653334806\\
9	0.000118828877883041\\
9.5	0.00011405178392903\\
10	0.000109705220394385\\
};

\addplot[area legend, draw=none, fill=mycolor5, fill opacity=0.25, forget plot]
table[row sep=crcr] {%
x	y\\
0	0.000725437577793409\\
0.5	0.000601721440593209\\
1	0.000490638029597816\\
1.5	0.000389936300465109\\
2	0.0003050217376549\\
2.5	0.000235149395043254\\
3	0.000194534713902167\\
3.5	0.000167936811129787\\
4	0.000154017334691074\\
4.5	0.000146597936990206\\
5	0.000143005775564507\\
5.5	0.000145251756858103\\
6	0.000143647638949042\\
6.5	0.000144779976483417\\
7	0.000138309700329276\\
7.5	0.00013617341351265\\
8	0.000131052752671894\\
8.5	0.000129679825024528\\
9	0.000126734020602021\\
9.5	0.000121352918164183\\
10	0.000117081841037608\\
10	0.000102348816065253\\
9.5	0.000105034050073943\\
9	0.000109782242923574\\
8.5	0.000116950001138581\\
8	0.000117292314542687\\
7.5	0.000123581590914427\\
7	0.000126428534435691\\
6.5	0.000130297098254039\\
6	0.000131842359628193\\
5.5	0.000136485556796872\\
5	0.00013578465779517\\
4.5	0.000139929651556551\\
4	0.000147325093079088\\
3.5	0.000159577011961972\\
3	0.000185591262216039\\
2.5	0.000223589907697235\\
2	0.0002908871137432\\
1.5	0.000376020373201006\\
1	0.000479778766201397\\
0.5	0.000591036059167179\\
0	0.000705775711420317\\
}--cycle;
\addplot [color=mycolor6]
  table[row sep=crcr]{%
0	0.0006736392740053\\
0.5	0.000574742620471505\\
1	0.00048508119445019\\
1.5	0.000407496388394507\\
2	0.000332369737738253\\
2.5	0.000271281204334376\\
3	0.000236815751699397\\
3.5	0.000215842191259404\\
4	0.000205192892018929\\
4.5	0.000197933359950487\\
5	0.000192571825729349\\
5.5	0.000192223147308951\\
6	0.000186515628033641\\
6.5	0.000184143993107157\\
7	0.000176793895856361\\
7.5	0.000171982525795441\\
8	0.000164118485467357\\
8.5	0.000159908806182419\\
9	0.000153426835283946\\
9.5	0.000148076642403203\\
10	0.000145393006918413\\
};

\addplot[area legend, draw=none, fill=mycolor6, fill opacity=0.25, forget plot]
table[row sep=crcr] {%
x	y\\
0	0.000688858285773845\\
0.5	0.000584674150818699\\
1	0.000494206009639818\\
1.5	0.000423601261406012\\
2	0.000344718992914941\\
2.5	0.000279920513759498\\
3	0.000242294855646427\\
3.5	0.000222956670326474\\
4	0.000212008920972275\\
4.5	0.000204562943421406\\
5	0.000198080407240535\\
5.5	0.000196458739824997\\
6	0.000190538345214627\\
6.5	0.000190774747697076\\
7	0.00018376299887858\\
7.5	0.000182875736206554\\
8	0.000176141831219002\\
8.5	0.000170622482168602\\
9	0.000159902770223344\\
9.5	0.000153810519975195\\
10	0.000153794260842122\\
10	0.000138088507040258\\
9.5	0.000140523910766654\\
9	0.000148914244403311\\
8.5	0.000151976169303982\\
8	0.000157456919645529\\
7.5	0.00016657130928628\\
7	0.000170518488787515\\
6.5	0.000179007073161538\\
6	0.000180407196088819\\
5.5	0.000187500106385556\\
5	0.000187184319738786\\
4.5	0.000192519492980864\\
4	0.000200997376907366\\
3.5	0.000211151796112957\\
3	0.000230975892566765\\
2.5	0.000265471521645288\\
2	0.000323839942035798\\
1.5	0.000396704349678827\\
1	0.000472519430710108\\
0.5	0.000561229284364776\\
0	0.000656690379486055\\
}--cycle;
\end{axis}
\end{tikzpicture}%
\hspace{0.2cm}
  \end{subfigure}
\caption{Prediction error and likelihood in five different driving scenarios.
}
\label{fig:eval_real_statistical}
\vspace{-10px}
\end{figure}

\mytodo{
FORWARD SIMULATION: CTRV vs. fully vs. map-based
specific example (one scene):
-s=0,n=high: forward simulation where fully outperforms simpler methods: RMSE-graph depending on prediction step
statistics (multiple scenes):
-s=0,n=high: RMSE-graph 
}

\mytodo{compare probabilistic trajectory (weighted with the route-combination probability) and with a uniformly distributed belief (equal weights) and with only the most likely trajectory}

\mytodo{RMSE for the different scenes with different models active; show that maneuver level is great!}

\mytodo{
Scene: intersection without traffic lights, one main road, one yielding road.
Situations part 1: ego on main, to be predicted on yielding to the left:
\begin{itemize}
\item route + curvature model (1 vehicle): distinguish between go-straight and turn maneuvers\\
curvature model leads to correct inference (e.g. actual left, inference  0/50/50 straight/right/left)
\item route + conservative conflict model (2 vehicles): distinguish between blocked lane and free lane\\
curvature model leads to wrong results (e.g. actual straight, inference 0/50/50)\\
conflict model leads to correct results (e.g. inference 50/0/50)
\item route + maneuver + conflict model (2 vehicles): distinguish between before and after\\
without maneuver using conservative conflict model leads to wrong results (e.g. actual straight before other, inference 0/100/0)\\
with maneuver using conflict model leads to correct results (e.g. inference 100/0/0)
\end{itemize}

maybe if we have time:
\begin{itemize}
\item predict both vehicles in aforementioned scenarios
\item more complicated scenes (e.g. with many cars)
\item roundabout
\end{itemize}}

\mytodo{add performance measure}

\balance
\section{Conclusions}
In this work, we proposed an interaction-aware prediction framework that is able to estimate route and maneuver intentions of drivers and predict complete scene developments in a combined fashion. 
Possible routes and maneuvers are generated online given a map and the current belief state.
The framework can handle a varying number of traffic participants and different road layouts without the need to predefine a discrete set of classes. 
It is capable of dealing with uncertainty in measurements and human behavior and interdependencies between drivers.
Its particle filtering nature allows to represent the non-linear system dynamics and the multi-modal and hybrid belief state.

Due to the combinatorial aspect of long-term motion prediction, the complexity of inference grows exponentially with the number of considered agents and possible intentions. Nevertheless, we show that in cases with close interaction between traffic participants, their interdependencies cannot be neglected.
Future work will focus on reducing this complexity and improving behavior model accuracy, e.g., by learning the action model from data, which in turn will reduce the number of needed particles. 
Furthermore, different possible plans of the ego vehicle could be taken into account within the forward simulation, in order to evaluate them with respect to how surrounding drivers are likely going to react.
Therefore, less conservative actions could be executed, respecting the influence of the ego vehicle on others.

\mytodo{can we describe in conclusion (instead of future work) that ego vehicle can be included??}
\mytodo{pros: forward simulation!! easy integration of ego vehicle action hypotheses to check how others would react!}
\mytodo{what are limitations? complexity! hand coded models bad for generalization!}

\mytodo{future work??
Future work will focus on both complexity and accuracy. Complexity reduction of the filtering framework could be achieved with Rao-Blackwellization, unscented Kalman filtering for a subset of states and factored particle filtering. \mytodo{cite??}
In order to increase prediction accuracy which also allows to reduce the number of needed particles, more accurate behavior models will be used by replacing the current hand-crafter models with models learned from data.
To further increase accuracy, reasonable priors for the possible intentions should be applied.
}

\bibliography{ICRA_2018}

\begin{thebibliography}{10}

\bibitem{lefevre_survey_2014}
S.~Lef{\`e}vre, D.~Vasquez, and C.~Laugier, ``A survey on motion prediction and
  risk assessment for intelligent vehicles,'' {\em Robomech J.}, vol.~1, no.~1,
  p.~1, 2014.

\bibitem{aoude_behavior_2011}
G.~S. Aoude, V.~R. Desaraju, L.~H. Stephens, and J.~P. How, ``Behavior
  classification algorithms at intersections and validation using naturalistic
  data,'' in {\em Intell. Veh. Symp. ({IV})}, pp.~601--606, {IEEE}, 2011.

\bibitem{barbier_classification_2017}
M.~Barbier, C.~Laugier, O.~Simonin, and J.~Iba{\~n}ez-Guzm{\'a}n,
  ``Classification of {Drivers} {Manoeuvre} for {Road} {Intersection}
  {Crossing} with {Synthetic} and {Real} {Data},'' in {\em Intell. Veh. Symp.
  ({IV})}, p.~7, {IEEE}, 2017.

\bibitem{phillips_generalizable_2017}
D.~J. Phillips, T.~A. Wheeler, and M.~J. Kochenderfer, ``Generalizable
  {Intention} {Prediction} of {Human} {Drivers} at {Intersections},'' in {\em
  Intell. Veh. Symp. ({IV})}, pp.~1665--1670, {IEEE}, 2017.

\bibitem{streubel_prediction_2014}
T.~Streubel and K.~H. Hoffmann, ``Prediction of driver intended path at
  intersections,'' in {\em Intell. Veh. Symp. ({IV})}, pp.~134--139, {IEEE},
  2014.

\bibitem{liebner_driver_2012}
M.~Liebner, M.~Baumann, F.~Klanner, and C.~Stiller, ``Driver intent inference
  at urban intersections using the intelligent driver model,'' in {\em Intell.
  Veh. Symp. ({IV})}, pp.~1162--1167, {IEEE}, 2012.

\bibitem{kumar_learning-based_2013}
P.~Kumar, M.~Perrollaz, S.~Lef{\`e}vre, and C.~Laugier, ``Learning-based
  approach for online lane change intention prediction,'' in {\em Intell. Veh.
  Symp. ({IV})}, pp.~797--802, {IEEE}, 2013.

\bibitem{bahram_combined_2016}
M.~Bahram, C.~Hubmann, A.~Lawitzky, M.~Aeberhard, and D.~Wollherr, ``A
  {Combined} {Model}- and {Learning}-{Based} {Framework} for
  {Interaction}-{Aware} {Maneuver} {Prediction},'' {\em IEEE Trans. Intell.
  Transp. Syst.}, vol.~17, pp.~1538--1550, June 2016.

\bibitem{klingelschmitt_combining_2014}
S.~Klingelschmitt, M.~Platho, H.-M. Gro{\ss}, V.~Willert, and J.~Eggert,
  ``Combining behavior and situation information for reliably estimating
  multiple intentions,'' in {\em Intell. Veh. Symp. ({IV})}, pp.~388--393,
  {IEEE}, 2014.

\bibitem{lefevre_risk_2012}
S.~Lef{\`e}vre, C.~Laugier, and J.~Iba{\~n}ez-Guzm{\'a}n, ``Risk assessment at
  road intersections: {Comparing} intention and expectation,'' in {\em Intell.
  Veh. Symp. ({IV})}, pp.~165--171, {IEEE}, 2012.

\bibitem{trautman_unfreezing_2010}
P.~Trautman and A.~Krause, ``Unfreezing the robot: {Navigation} in dense,
  interacting crowds,'' in {\em Int. Conf. Intell. Robot. and Syst. (IROS)},
  pp.~797--803, IEEE, 2010.

\bibitem{kuhnt_understanding_2016}
F.~Kuhnt, J.~Schulz, T.~Schamm, and J.~M. Z{\"o}llner, ``Understanding
  interactions between traffic participants based on learned behaviors,'' in
  {\em Intell. Veh. Symp. ({IV})}, pp.~1271--1278, {IEEE}, 2016.

\bibitem{tran_online_2014}
Q.~Tran and J.~Firl, ``Online maneuver recognition and multimodal trajectory
  prediction for intersection assistance using non-parametric regression,'' in
  {\em Intell. Veh. Symp. ({IV})}, pp.~918--923, {IEEE}, 2014.

\bibitem{armand_modelling_2013}
A.~Armand, D.~Filliat, and J.~Iba{\~n}ez-Guzm{\'a}n, ``Modelling stop
  intersection approaches using gaussian processes,'' in {\em Int. Conf.
  Intell. Transp. Syst. ({ITSC})}, pp.~1650--1655, {IEEE}, 2013.

\bibitem{gindele_learning_2013}
T.~Gindele, S.~Brechtel, and R.~Dillmann, ``Learning context sensitive behavior
  models from observations for predicting traffic situations,'' in {\em Int.
  Conf. Intell. Transp. Syst. ({ITSC})}, pp.~1764--1771, {IEEE}, 2013.

\bibitem{lenz_deep_2017}
D.~Lenz, F.~Diehl, M.~T. Le, and A.~Knoll, ``Deep neural networks for
  {Markovian} interactive scene prediction in highway scenarios,'' in {\em
  Intell. {Veh.} {Symp.} ({IV}), 2017 {IEEE}}, pp.~685--692, IEEE, 2017.

\bibitem{schulz_estimation_2017}
J.~Schulz, K.~Hirsenkorn, J.~L\"{o}chner, M.~Werling, and D.~Burschka,
  ``Estimation of collective maneuvers through cooperative multi-agent
  planning,'' in {\em Intell. Veh. Symp. ({IV})}, pp.~624--631, {IEEE}, 2017.

\bibitem{wheeler_analysis_2016}
T.~A. Wheeler, P.~Robbel, and M.~J. Kochenderfer, ``Analysis of microscopic
  behavior models for probabilistic modeling of driver behavior,'' in {\em Int.
  Conf. Intell. Transp. Syst. ({ITSC})}, pp.~1604--1609, {IEEE}, 2016.

\bibitem{platho_predicting_2013}
M.~Platho, H.-M. Gro{\ss}, and J.~Eggert, ``Predicting velocity profiles of
  road users at intersections using configurations,'' in {\em Intell. Veh.
  Symp. ({IV})}, pp.~945--951, {IEEE}, 2013.

\bibitem{treiber_Congested_2000}
M.~Treiber, A.~Hennecke, and D.~Helbing, ``Congested {Traffic} {States} in
  {Empirical} observations and {Microscopic} {Simulations},'' {\em Phys. Rev.
  E}, vol.~62, pp.~1805--1824, Aug 2000.

\bibitem{schubert_comparison_2008}
R.~Schubert, E.~Richter, and G.~Wanielik, ``Comparison and evaluation of
  advanced motion models for vehicle tracking,'' in {\em Int. Conf. Inform.
  Fusion}, pp.~1--6, {IEEE}, 2008.

\end{thebibliography}
\bibliographystyle{ieeetr}

\clearpage

\end{document}